\documentclass[10pt,journal,cspaper,compsoc]{IEEEtran}
\usepackage{CJKutf8}	
	
\usepackage{subcaption}
\usepackage{color}

\usepackage{booktabs}
\usepackage{array}
\usepackage{url}
\usepackage{epsfig}
\usepackage{graphicx}
\usepackage{amsmath}
\usepackage{amssymb}
\usepackage{color} 
\usepackage{marvosym}
\usepackage{bbm}
\usepackage{cite}
\usepackage[ruled,lined,linesnumbered]{algorithm2e}
\usepackage{enumerate}

\usepackage{subcaption}

\usepackage{algorithmic}

\usepackage{url}
\usepackage{multirow}
\usepackage[usenames,dvipsnames,table]{xcolor}

\usepackage[table]{xcolor}
\usepackage{pgfplotstable}
\usepackage{colortbl}
\usepackage{pgfplots}
\pgfplotstableset{
	/color cells/min/.initial=0,
	/color cells/max/.initial=200,
	/color cells/textcolor/.initial=,
	columns/x/.style={column name={From$\setminus$To}, postproc cell content/.code={}},
	columns/1/.style={column name={0$\sim$10}},
	columns/2/.style={column name={11$\sim$18}},
	columns/3/.style={column name={19$\sim$30}},
	columns/4/.style={column name={31$\sim$40}},
	columns/5/.style={column name={41$\sim$50}},
	columns/6/.style={column name={51$\sim$60}},
	columns/7/.style={column name={60+}},
	every row 0 column 0/.style={postproc cell content/.style={@cell content={0$\sim$10}}},
	every row 1 column 0/.style={postproc cell content/.style={@cell content={11$\sim$18}}},
	every row 2 column 0/.style={postproc cell content/.style={@cell content={19$\sim$30}}},
	every row 3 column 0/.style={postproc cell content/.style={@cell content={31$\sim$40}}},
	every row 4 column 0/.style={postproc cell content/.style={@cell content={41$\sim$50}}},
	every row 5 column 0/.style={postproc cell content/.style={@cell content={51$\sim$60}}},
	every row 6 column 0/.style={postproc cell content/.style={@cell content={60+}}},
	color cells/.code={%
		\pgfqkeys{/color cells}{#1}%
		\pgfkeysalso{%
			postproc cell content/.code={%
				\begingroup
				\pgfkeysgetvalue{/pgfplots/table/@preprocessed cell content}\value
				\ifx\value\empty
				\endgroup
				\else
				\pgfmathfloatparsenumber{\value}%
				\pgfmathfloattofixed{\pgfmathresult}%
				\let\value=\pgfmathresult
				\pgfplotscolormapaccess
				[\pgfkeysvalueof{/color cells/min}:\pgfkeysvalueof{/color cells/max}]%
				{\value}%
				{\pgfkeysvalueof{/pgfplots/colormap name}}%
				\pgfkeysgetvalue{/pgfplots/table/@cell content}\typesetvalue
				\pgfkeysgetvalue{/color cells/textcolor}\textcolorvalue
				\toks0=\expandafter{\typesetvalue}%
				\xdef\temp{%
					\noexpand\pgfkeysalso{%
						@cell content={%
							\noexpand\cellcolor[rgb]{\pgfmathresult}%
							\noexpand\definecolor{mapped color}{rgb}{\pgfmathresult}%
							\ifx\textcolorvalue\empty
							\else
							\noexpand\color{\textcolorvalue}%
							\fi
							\the\toks0 %
						}%
					}%
				}%
				\endgroup
				\temp
				\fi
			}%
		}%
	}
}

\begin{document}

\title {Human-centric Relation Segmentation: Dataset and Solution}

\author{Si Liu, Zitian Wang, Yulu Gao, Lejian Ren, Yue Liao, Guanghui Ren, Bo Li, Shuicheng Yan
\IEEEcompsocitemizethanks{\IEEEcompsocthanksitem  Si Liu, Zitian Wang, Yulu Gao,  Yue Liao  and Bo Li are with Beihang University. Lejian Ren, Guanghui Ren are with Chinease Academy of Science. Shuicheng Yan is with Sea AI Lab (SAIL).}
}


\markboth{IEEE TRANSACTIONS ON PATTERN ANALYSIS AND MACHINE INTELLIGENCE, VOL. XX, NO. X, X 20XX}
{Shell \MakeLowercase{\textit{et al.}}: Bare Demo of IEEEtran.cls for Computer Society Journals}

\IEEEcompsoctitleabstractindextext{

\begin{abstract}
Vision and language understanding techniques have achieved remarkable progress, but currently  it is still difficult to well handle problems involving very fine-grained details.
For example,  
when the robot is told to ``bring me the book in the girl's left hand'', most existing methods would fail  if the girl  holds one book respectively in her left and right hand.
In this work, we introduce a new task named human-centric relation segmentation (HRS), \textcolor{black}{as a fine-grained case of HOI-det.}
HRS aims to predict the relations between the human and surrounding entities and identify the  relation-correlated human parts, which are represented as pixel-level masks.
For the above exemplar case, our HRS  task produces results in the form of relation triplets  $\langle$girl [left hand], hold, book$\rangle$ and exacts segmentation masks of the book, with which the robot can easily accomplish the grabbing task.
Correspondingly, we collect a new Person In Context (PIC) dataset for this new task, which contains $17,122$ high-resolution images and densely annotated entity segmentation and relations, including $141$ object categories, $23$ relation categories and $25$ semantic human parts.
We also propose a Simultaneous Matching and Segmentation (SMS) framework as a solution to the HRS task.
It contains three parallel branches for entity segmentation, subject object matching and human parsing respectively.
Specifically, the entity segmentation branch obtains entity masks by dynamically-generated conditional convolutions; the subject object matching branch detects the existence of any relations, links the corresponding subjects and objects by displacement estimation and classifies the interacted human parts; and the human parsing branch generates the pixelwise human part labels.
Outputs of the three branches are fused to produce the final HRS results.
Extensive experiments on PIC and V-COCO datasets show that the proposed SMS method outperforms baselines with the 36 FPS inference speed.
Notably,  SMS outperforms the best performing baseline $m$-KERN with only $17.6\%$ time cost.
The dataset and code will be released at \url{http://picdataset.com/challenge/index/}.

\end{abstract}


}

\maketitle

\newcommand{\etal}{\textit{et al}.}
\newcommand{\ie}{\textit{i}.\textit{e}.}
\newcommand{\eg}{\textit{e}.\textit{g}.}

\section{Introduction} \label{sec:intro}
\begin{figure*}[t]
	\centering
	\includegraphics[width=.9\textwidth]{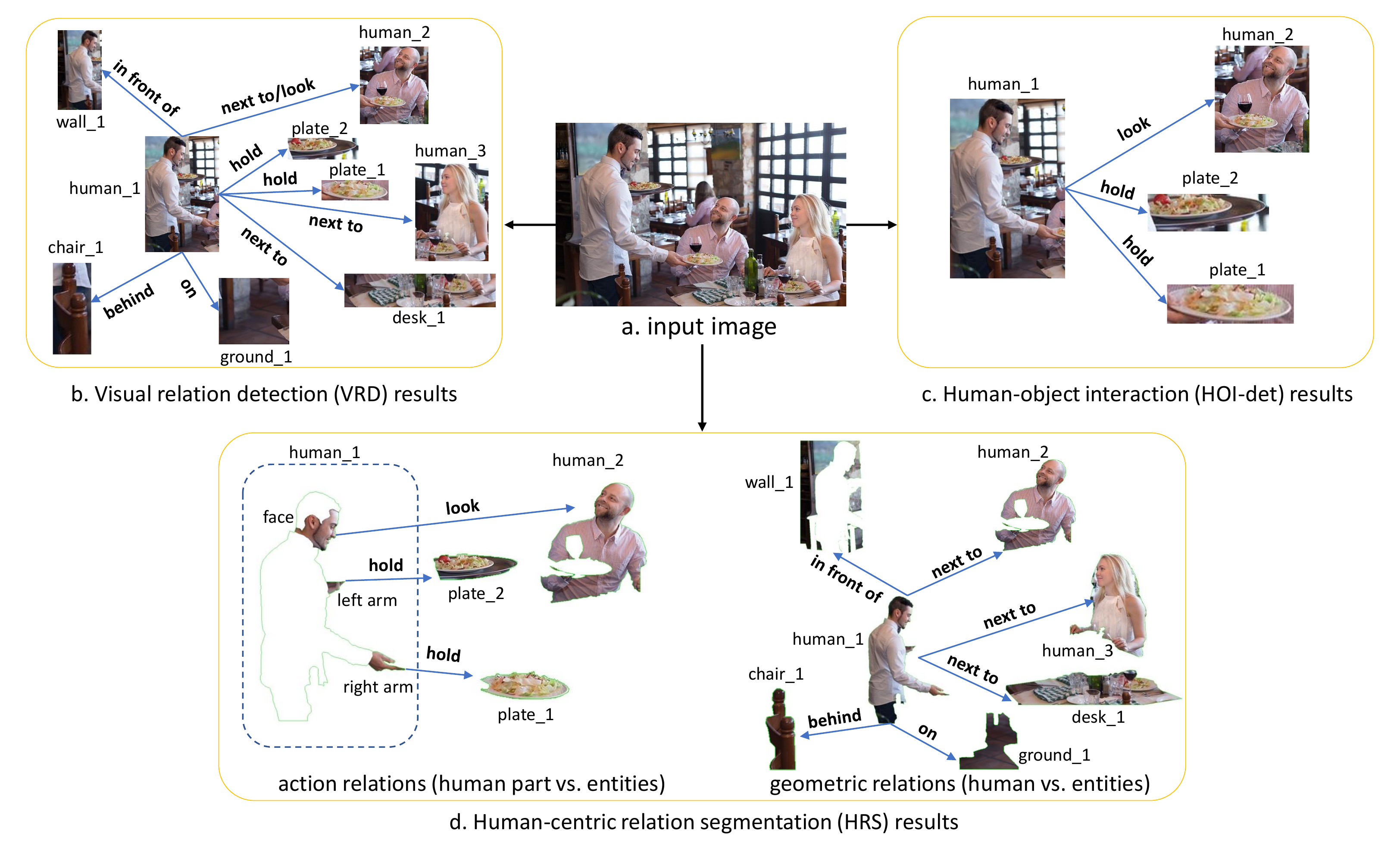}
	\caption{Comparison of the proposed HRS task with VRD and HOI-det.  }
	\label{fig:main}
\end{figure*}

\IEEEPARstart{R}ecently, great progress has been made in the vision \&  language understanding areas such as  embodied AI~\cite{habitat2020sim2real,das2018embodied,anderson2018vision,qi2020reverie},
referring expression/segmentation~\cite{lu2019vilbert,chen2019uniter,su2019vl},
visual question answering~\cite{antol2015vqa,anderson2018bottom}  and  text guided image editing~\cite{li2019manigan}, etc.
However in some scenarios involving understanding very fine-grained details, existing algorithms often give unsatisfactory results and human part level  relation segmentation is required.
For example, in a remote embodied visual referring expression task~\cite{qi2020reverie},
given the image in Figure \ref{fig:main}a and the instruction ``pass me the plate the waiter holds in his left  arm'', existing algorithms would fail to distinguish the two plates held by the waiter.
For such a case, we argue that inferring the part level relation triplet   $\langle$human\_1 [left arm], hold, plate\_2$\rangle$ (shown in Figure \ref{fig:main}d)  is very helpful for  identifying which plate is the  target one, and moreover, the segmentation mask of the plate also need be estimated to facilitate the robot grabbing.
For another example, in the text guided image editing task \cite{li2019manigan}, given  the text instruction ``erase the plate held by the man's right arm'', not only the fine-grained relation  $\langle$human\_2 [right arm], hold, plate\_1$\rangle$ but also the precise pixel-level mask of the plate is needed.

Targeted at problems that require fine-grained details,  we propose a new task --- human-centric relation segmentation (HRS), \textcolor{black}{which is a fine-grained case of human-object interaction detection (HOI-det).} HRS aims to predict the relations between human and its surrounding entities and identify the  relation-correlated human parts, which are  represented as pixel-level masks. In this paper, we use \emph{entity} to include \emph{thing} and \emph{stuff}.

This new HRS task is partially related to Visual Relation Detection (VRD) \cite{krishna2017visual,zellers2018neural,zhang2017visual,yang2018graph,tang2020unbiased} which aims at understanding the relations between entities in the image, and \textcolor{black}{HOI-det}  \cite{gupta2015visual,chao2018learning} which tries to estimate the relations between human and surrounding objects.
An illustration of VRD, HOI-det and HRS  is shown in Figure \ref{fig:main}.
Given an input image in Figure \ref{fig:main}a, VRD generates several $\langle$subject, relation, object$\rangle$ triplets, such as $\langle$human\_1, next to, desk\_1$\rangle$, as shown in Figure \ref{fig:main}b, while  HOI-det outputs $\langle$human, action, object$\rangle$ triplets like $\langle$human\_1, hold, plate\_1$\rangle$, which  focuses on the relations between human and objects it interacts with, as shown in Figure \ref{fig:main}c.
Both VRD and HOI-det adopt bounding boxes to represent the subject and object  regarding a certain relation.
By contrast, our HRS task adopts pixel-level masks to represent both subject and object, and divides relations into two different types, as shown in Figure \ref{fig:main}d.
The first type of relation is called \emph{action relation}  defined between specific human parts and the corresponding entity, e.g.    $\langle$human\_1 [right arm], hold, plate\_1$\rangle$. The subjects corresponding to such relations are human parts correlated to action rather than the whole human body.
The other type is \emph{geometric relation} defined between human and the corresponding entity that describes relative positions, like $\langle$human\_1, behind, chair\_1$\rangle$.

\begin{table}[h]
	\caption{Comparison of VRD, HOI-det and HRS. Notations ``part'',``geo'' and ``seg'' denote ``human part'',   ``geometric'' and  ``segmentation'' respectively.}
	\begin{center}
		\begin{tabular}{ c|c|c|c|c}
			Task  & Subject & Relation & Object & Representation\\
			\hline
			VRD  &entity & action+geo& entity &box\\
			\hline
			HOI-det    &human  &action &thing   &box \\
			\hline
			HRS      &human+part  & action+geo &entity  & seg  \\
			\hline
		\end{tabular}
	\end{center}
	\label{tb:task_compare}
\end{table}

Compared with VRD and HOI-det,
HRS can capture not only the relative position information, but also the detailed action information.
Through the classification of interacted human parts , HRS can distinguish $\langle$human [left arm], throw, ball$\rangle$ and $\langle$human [both arms], throw, ball$\rangle$.
The HRS adopts the more expressive mask  representation rather than bounding boxes, which can describe exact entity shapes and  contain little background area and thus is fairly beneficial to downstream tasks.
Also,  the masks are more suitable for representing stuff, such as  sky or ocean, which has no fixed shape and is often non-convex.
A summary of comparisons among VRD, HOI-det and the proposed HRS is given in Table \ref{tb:task_compare}.

Since existing datasets either have no part level relation annotations or no mask segment annotations,  we collect and annotate a large high-resolution dataset, called Person in Context (PIC), to  facilitate research on the new HRS task.
The PIC dataset contains $17,122$ images  with both segmentation  and relation labels.
For segmentation, we densely label $141$ kinds of thing and stuff entities.
We also label $25$ human parts and $16$ human key points.
For relation, we label $14$ action relations 
and $9$ geometric relations.

In addition, we propose a Simultaneous Matching  and Segmentation (SMS) framework to solve the new HRS task, with compact structure and simple design.
SMS contains three parallel branches.
1) An \emph{entity segmentation} branch segments entities in the  image by identifying their centers and corresponding masks.
To detect the entity centers,  we estimate entity center  heatmaps  whose highest responses correspond to the locations and categories of the entity centers.
To generate the masks,  we produce  mask kernels and mask features in parallel.
Depending on the location of the detected entity center, a specific location-aware mask kernel  is dynamically  generated  and convoluted with the mask feature to generate the mask of the entity.
2) A \emph{subject object matching} branch pairs the human  with its interacted entities and  identifies their relations.
Concretely, we detect the relation point (the middle point of the subject and object), and also estimate the displacements between each relation point  and the subject/object as well as classify the human parts involved in the interaction.
The subject and object linked by the same relation point are paired and form a  triplet.
3) A \emph{human parsing} branch produces semantic human parsing.
The  results produced by these three branches are  fused to generate the final HRS result.
SMS is able to achieve $36$ FPS for $512\times 512$ sized images in the $\text{Recall}@25$ setting.
In addition, we extensively benchmark the dataset with the baselines modified from state-of-the-art VRD and HOI-det methods.

Our contributions can be summarized as follows.
1) We propose a new human-centric relation segmentation task, which can benefit a lot of cognitive-like tasks. 
\textcolor{black}{HRS is  a fine-grained case of HOI-det.}
2) We collect, annotate  and release a large high-resolution Person in Context (PIC) dataset for facilitating exploring the task.
3) we propose an one-stage Simultaneous Matching  and Segmentation (SMS) framework, which integrates the entity segmentation and human-centric relation prediction in an unified  model, achieving competitive performance at real time inference speed.  
Both the PIC dataset and the proposed SMS method will be released at http://picdataset.com/challenge/index/.

\begin{figure*}[t]
	\centering
	\includegraphics[width=\textwidth]{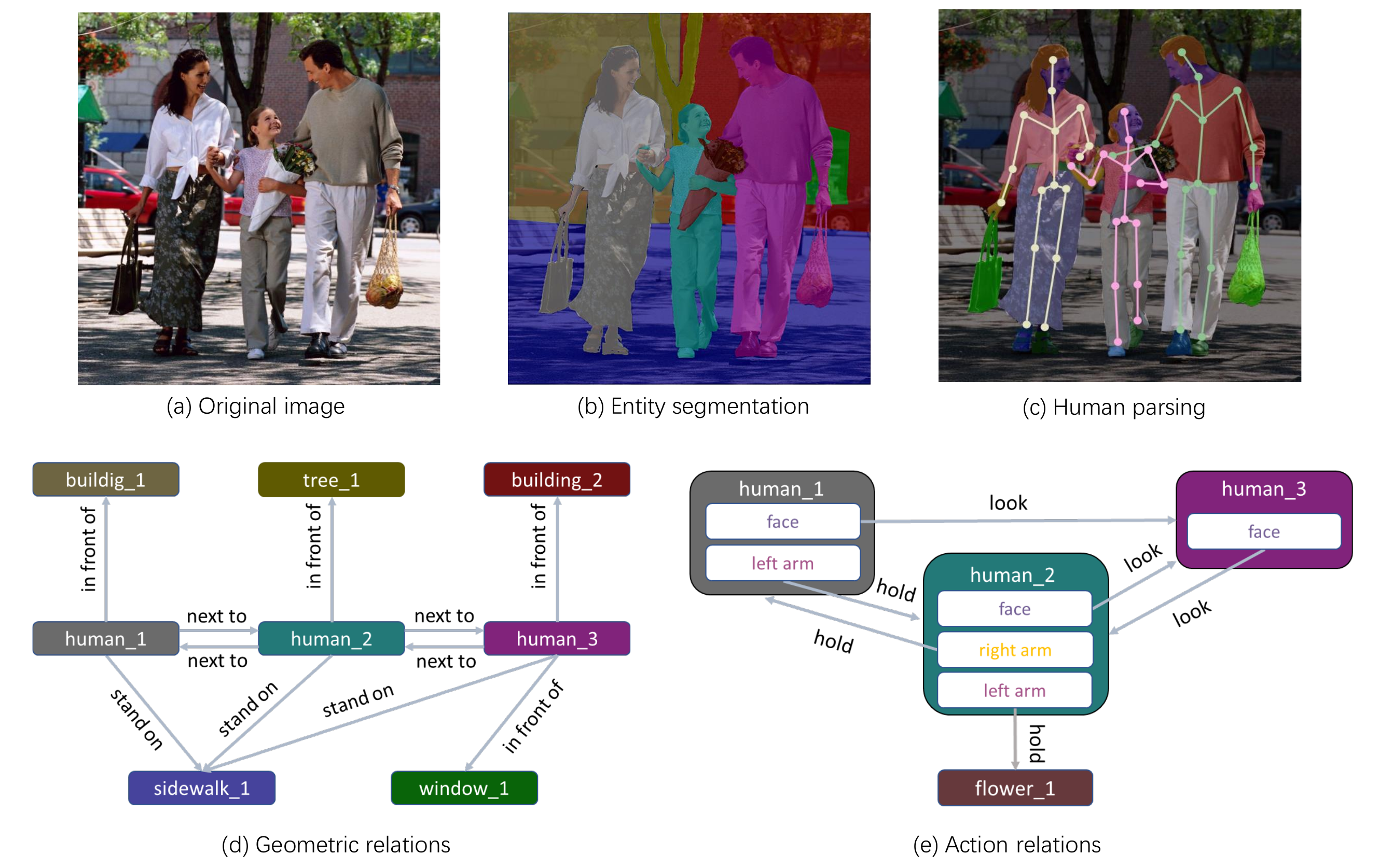}
	\caption{An example of the PIC dataset. For an image (a), both the entity segmentation (b) and human parsing (c) are labelled.
	We also label geometric and action relations. 
	In the geometric relation graph (d), the nodes are entity instances of (b) while the edges are  spatial relations.
    In the  action relation graph, there are two kinds of nodes. The non-human nodes are also entity instances of (b) while the human nodes contain the interacted human parts in (c) which make the action. }
	\label{fig:final-dataset}
\end{figure*}

\section{Related Work} \label{sec:relatedworks}

\begin{figure}[h]
	\centering
	\includegraphics[width=0.45\textwidth]{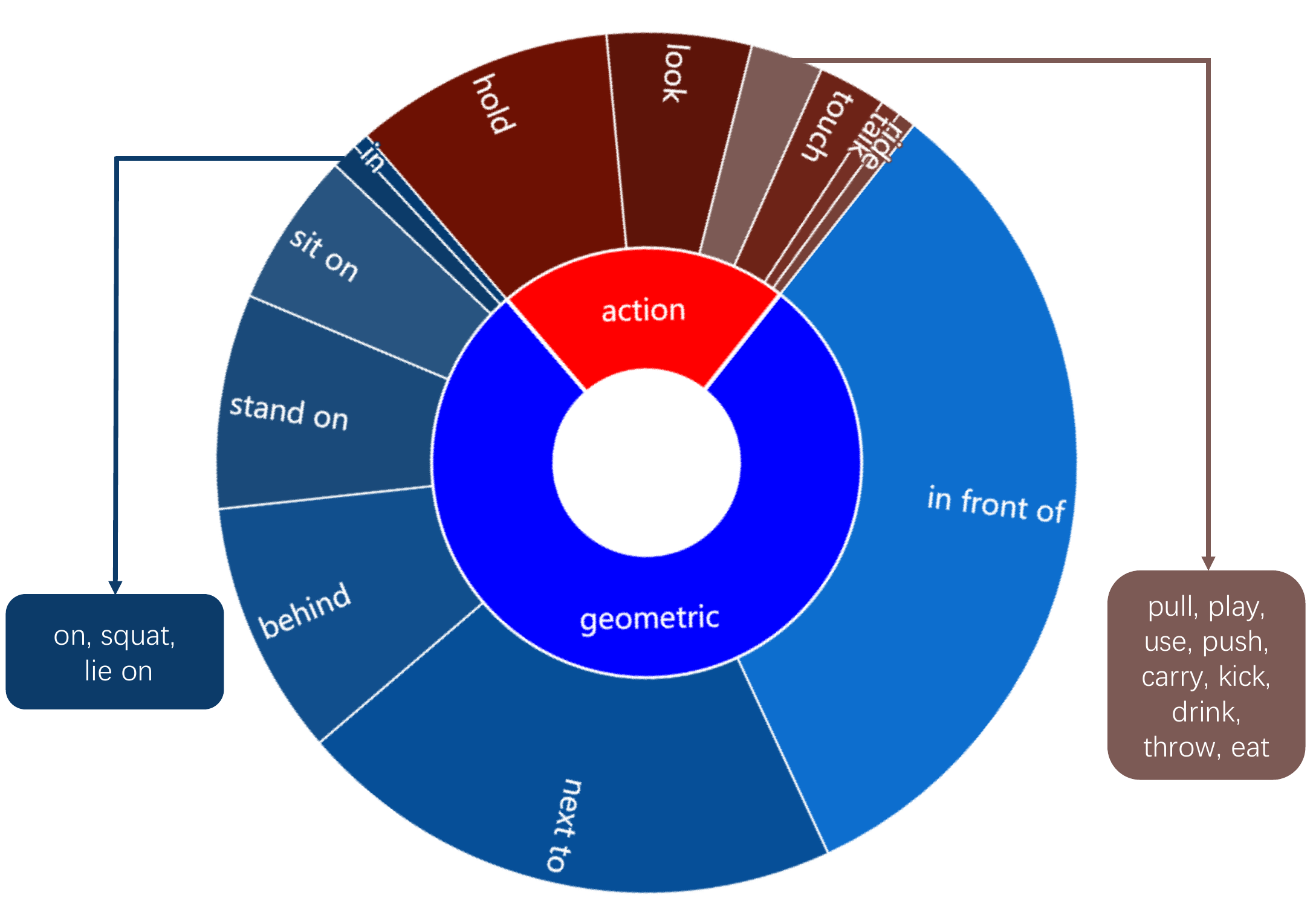}
	\caption{Distribution of action  and geometric relations.}
	\label{fig:predicate}
\end{figure}

\subsection{VRD and HOI-det}
\emph{VRD} aims to detect the relation triplets composed of subject, relationship and object in the input image.
Most existing VRD methods follow a two-step pipeline including  object detection  and  relation estimation.
Existing research mostly focuses on the second step. Lu et al. proposed a two-way framework that finetunes the likelihood of a predicted relationship using language priors from semantic word embeddings \cite{lu2016visual}.
Xu et al. introduced a method to pass message between objects and relations to improve triplet predictions \cite{xu2017scene,zellers2018neural,li2017vip}.  Zhang  et al. attempted to learn the relation in a low-dimensional space \cite{zhang2017visual}. Recently, the graph convolution \cite{kipf2016semi}  is used to gather the information from different nodes \cite{yang2018graph}.

\begin{figure*}[h]
	\centering
	\includegraphics[width=0.9\textwidth]{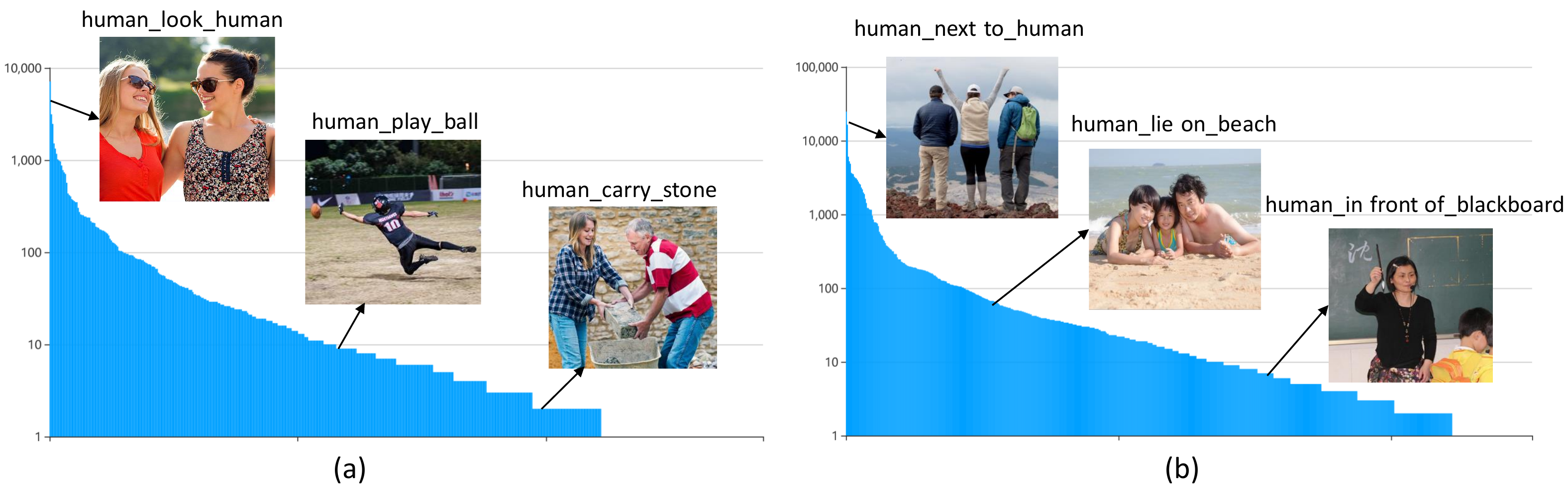}
	\caption{(a) Distribution of action triplets and (b) distribution of geometric triplets.}
	\label{fig:all_triplet}
\end{figure*}

\emph{HOI-det} as a special case of the VRD task, aims to detect the action relations with subjects specified to be ``human''.
HOI-det  needs to estimate the bounding boxes of human and object as well as their relations, e.g.,  actions and  geometric relations.
Most previous methods follow a two-stage pipeline, first detecting all candidate subjects and objects and then predicting their interactions.
Pang et al. proposed a contextual attention framework to exploit the rich context information that is crucial for HOI prediction \cite{wang2019deep}. GPNN uses a message passing mechanism to reason upon graph structured information \cite{qi2018learning}.
Feng et al. proposed a turbo learning method which views human pose and HOI as complementary information to each other and optimize both tasks in an iterative manner \cite{feng2019turbo}.
Our proposed \emph{HRS} explores the geometric relations and action relations between humans and entities. Moreover, it requires fine-grained classification on the human parts that make the actions.
We propose a one-stage method that  contains parallel branches, the intermediate results of which can be fused to form HRS results.

Both \cite{wang2020learning}   and our method  detect the relations by combining relation point detection and  interaction grouping. Compared with traditional HOI-det methods, both \cite{wang2020learning}   and our method  directly detect relation from the whole feature map, thus utilizing richer context  and reducing the computational cost that originally scales quadratically with the number of entities. 
The differences between~\cite{wang2020learning} and our method lie in the following  aspects.
	($\romannumeral1$) The two methods are designed for {different tasks}. ~\cite{wang2020learning} tackles HOI-det, while we tackle HRS which provides fine-grained understanding of human and the surrounding environment.
	($\romannumeral2$) ~\cite{wang2020learning} adopts an {extra} and {independent} detector  to detect the entities, while our method integrates entity segmentation into the {unified} SMS framework. Thus our method is more compact and  fast during inference. 
	($\romannumeral3$) Both methods contain similar  {keypoint-based} relation point detection modules. But their  object/entity detection modules are different.  Specifically,  \cite{wang2020learning} adopts {anchor-based} Faster R-CNN \cite{ren2016faster} as the  object detector, which  is {inconsistent} with the keypoint-based relation point detection module.  But in our method, the entity detection is conducted in a {keypoint-based}  manner. Thus both our entity detection module and relation point detection module are {keypoint-based} and thus {consistent},
which 
enables the latter to benefit from the former.
($\romannumeral4$) ~\cite{wang2020learning} defines the interaction point as the midpoint between the {bounding box center} of subject and object, while our method defines the relation point as the midpoint between the {mass center} of subject and object  segments, which is tailored for our relation segmentation task.

\subsection{Instance Segmentation and Relation Segmentation}
Instance segmentation aims to detect the instances in the image and represent them by pixel-level masks.
We follow the instance segmentation methods to obtain the masks for subjects and objects of detected relations.
Instance segmentation methods can be roughly divided into two kinds.
First,  detection-based methods achieve successful segmentation relying on the object proposals \cite{dai2016instance,dai2016instance,liu2018path}.
Li et al. proposed a FCN that adopts position-sensitive score maps to detect and segment objects simultaneously \cite{li2017fully}.
Mask R-CNN \cite{he2017mask} is built on the Faster R-CNN and uses an additional mask branch to generate object masks.
Chen et al. introduced a cascade architecture to interweave detection and segmentation, and progressively refine both results \cite{chen2019hybrid}.
Fu et al. built an instance segmentation model on single-shot detectors, retaining fast inference speed and showing competitive accuracy \cite{fu2019retinamask}.
Second, detection-free methods where the instance masks do not rely on the pre-detected bounding boxes are also very popular.
For example, some metric learning based methods \cite{newell2017associative,fathi2017semantic,neven2019instance}  learn an embedding for each pixel and cluster the pixels that are close in the embedding space to generate instance masks.
Recently, Polarmask \cite{xie2019polarmask} represents the instance masks by discrete contour points and directly regresses the  contour points. SOLO \cite{wang2019solo} assigns instance masks to specific instance categories according to locations and scales, converting instance masks prediction to a simpler classification problem.
SOLOv2 \cite{wang2020solov2} learns adaptive mask kernels and mask features to generate the mask for each instance conditioned on the instance location. Our method follows the SOLOv2 mechanism to get the mask of both human and surrounding entities (thing and stuff)  to achieve accuracy-speed trade-off.

~\cite{zhou2020cascaded} proposes a multi-stage network which progressively refines the instance segmentation and relation prediction according to the output of its previous stage. The differences between~\cite{zhou2020cascaded} and our method is  that they build the instance localization network and an interaction recognition network work in a \emph{cascade} manner. More specifically, an instance localization network first generates human and object proposals, and then an instance recognition network  identifies the action  for each human-object pair. Thus their method suffers from large computational cost and prediction latency. To the contrary, our method establishes a \emph{simultaneous} matching and segmentation framework, where both the entities and their relations are estimated by keypoints heatmap in the one-stage model with much less complexity.

\subsection{Human Parsing}
In HRS setting, the subject of one action relation is set as the relation-correlated human parts. Thus HRS requires human part segments which can be obtained through human parsing.
Human parsing has received increasing interest over recent years  \cite{chen2014detect,liang2015deep,liang2015human}. Liang et al. introduced a large-scale dataset LIP and proposed a self-supervised structure-sensitive learning approach \cite{liang2018look}.
Gong et al. collected a multi-human parsing dataset and  proposed a detection-free model \cite{gong2018instance}.
 Zhao et al. presented a fine-grained multi-human parsing  dataset MHPv2 and  proposed a nested adversarial network model  \cite{zhao2018understanding}.
 In our HRS task, we  estimate the actions between human parts and surrounding entities

\begin{table}
	\centering
	\begin{subtable}[t]{3.2in}
		\centering
	\begin{tabular}{|c|c|c|c|c|}
\hline
face  & hair & coat & uppercloth & left\_arm      \\
\hline
right\_arm &	pants & left\_leg &right\_leg  & skin      \\
\hline
beard&	headwear    & jewelry&dress    &skirt  \\
\hline
left\_shoe&	right\_shoe    & bag&tie    &hat  \\
\hline
scarf &	sunglasses    & gloves&socks    &belt  \\
\hline
		\end{tabular}
		\caption{25 Human parsing labels.}\label{table:parsing}
	\end{subtable}
	\quad
	\begin{subtable}[t]{3.2in}
		\centering
	\begin{tabular}{|c|c|c|c|}
				\hline
			head\_top  & upper\_neck & thorax & left\_shoulder   \\
\hline
right\_shoulder  &  			left\_elbow &	right\_elbow & left\_wrist       \\
\hline
right\_wrist  & pelvis &			left\_hip&	right\_hip     \\
\hline
left\_knee &right\_knee    &left\_ankle &right\_ankle\\
\hline
		\end{tabular}
		\caption{16 Human pose labels.}\label{table:pose}
	\end{subtable}
	\caption{Human parsing and pose labels.}\label{table:paring_pose}
\end{table}

\section{Person In Context Dataset} \label{sec:dataset}

In this section, we introduce the collected PIC dataset for the HRS task.
We have spent nearly three years to continuously refine the dataset.
Notably, we organized ECCV 2018 the 1st PIC workshop/challenge and ICCV 2019 the 2nd PIC  workshop/challenge, which attracted worldwide competitors.
The feedbacks from the participants have also been explored to help refine the dataset.
In this paper we provide the latest version (a.k.a PIC 3.0) which is different from that used in ECCV 18 and ICCV 19 challenges.
In future, we will keep refining it by collecting more images with strict copyright verification and labeling more diverse relations.

\subsection{Data Collection}

The data collection is done with three steps.
 \begin{enumerate}
 	\item Data crawling: PIC contains both indoor and outdoor images, which are crawled from Flickr website with copyrights. 
 	Query words for retrieving indoor pictures include cook, party, drink, watch tv,  eat, etc., and those for outdoor ones include play ball, run, ride, outing, picnic, etc.
 	In this way, the collected data enjoy great diversities in terms of scenario, appearance, viewpoint, light condition and occlusion.
 	\item Data filtering: We filter out the images with low resolution or without human.
 	Then we  calculate the distributions of the relations.
 	We note that the relation distribution shows a long tail.
 	\item Data balancing: We recollect the data for relations with lower frequency to balance the data distribution. Therefore, the long tail effect  is alleviated to some extent.
 \end{enumerate}

\begin{table*}[htb]
	\caption{Comparison with existing datasets. }
	\begin{center}
		\begin{tabular}{c|c|c|c|c|c|c}
			Dataset  & Stuff & Segmentation & Relation & Human part & Human Pose & Average resolution\\
			\hline
			Visual Gennome\cite{krishna2017visual}  & \checkmark   &  $\checkmark$    &\checkmark   &  $\times$ &  $\checkmark$ & 413*500 \\
			\hline
			VRD\cite{lu2016visual}  &\checkmark    & $\times$    &\checkmark & $\times$ & $\times$ & 764*950 \\
			\hline
			VCOCO \cite{gupta2015visual}   & $\times$    &$\checkmark$   &\checkmark  &$\times$ & $\checkmark$ & 481*578 \\
			\hline
			HICO-DET   \cite{chao2018learning}   & $\times$    &$\times$   &\checkmark  & $\times$ & $\times$ & 497*593 \\
			\hline
			PIC      &\checkmark   &\checkmark   &\checkmark  & \checkmark& \checkmark & 1427*1882\\
			\hline
		\end{tabular}
	\end{center}
	\label{tb:dataset_stat}
\end{table*}

\subsection{Data Annotation}
We sequentially label the entity segmentation and relations.
For ``human'' entities, we densely annotate the  human part segments and human keypoints.
Two kinds of relations (action and geometric relations) are annotated between humans and surrounding entities.
The detailed annotation procedure is as follows:

\textbf{Entity segmentation:}
We first annotate $141$ kinds of things and stuff in the images. The entity categories cover a wide range of indoor and outdoor scenes, including office, restaurant, seaside, snowfield, etc.
For each entity falling into predefined categories, we label it with its class and pixel-level mask segment. The disconnected regions of stuff are viewed as different entities.
Some examples are shown in Figure \ref{fig:final-dataset}b.

\textbf{Human parts \& human keypoints:}
Human part segments and human keypoints are annotated for all the ``human'' entities.
Our dataset also provides human keypoints annotation in order to facilitate the follow-up study.
As shown in Table \ref{table:parsing} and Table \ref{table:pose},   we annotate $25$ different human parts inspired by ATR dataset \cite{liang2015deep}.
As shown in Table \ref{table:pose}, we annotate  $16$ different human keypoints following the definition of  \cite{zhao2018understanding,li2017multiple}. Some human pose and parsing examples are shown in Figure \ref{fig:final-dataset}c.

\textbf{Relation:}
We annotate relations between the annotated ``human'' entities and their surrounding entities. We define 14 kinds of geometric relations and 9 kinds of action relations which are common in daily expressions.

\textit{Geometric relations} are defined upon the human and surrounding context.   Some examples are shown in Figure \ref{fig:final-dataset}d, such as
 $\langle$human\_2, next to, human\_3$\rangle$ and  $\langle$human\_2, in front of, tree\_1$\rangle$.

\textit{Action relations} are defined between the human part and the interacted entities.
Some action relations are shown in Figure \ref{fig:final-dataset}e, like $\langle$human\_2 [left arm], hold, flower\_1$\rangle$ and  $\langle$human\_1 [left arm], hold, human\_2$\rangle$, etc.

Note that there may exist multiple relations between different human parts and an object. For example, $\langle$human [face], look at, ball$\rangle$, $\langle$human [left arm], hold, ball$\rangle$ and $\langle$human [right arm], hold, ball$\rangle$ may co-exist.
Moreover, there may even exist multiple relations between one human part and one object. For example, $\langle$human [left arm], hold, phone$\rangle$ and  $\langle$human [left arm], use, phone$\rangle$ may co-exist.
Besides, geometric relations are not exclusive either. For example,  $\langle$human\_1, next to, human\_2$\rangle$ and $\langle$human\_1, in front of, human\_2$\rangle$   may co-exist.

\subsection{Dataset Statistics}

The PIC dataset contains $17,122$  images with $141$ categories of entity segments, $23$  kinds of relations between human and entities. Each human is annotated with  $25$ different human parts and $16$ human keypoints.
PIC contains $8.1$ entities (including $3.8$ human entities) and $10.4$ relations per image on average.

\begin{figure}[h]
	\centering
	\includegraphics[width=0.45\textwidth]{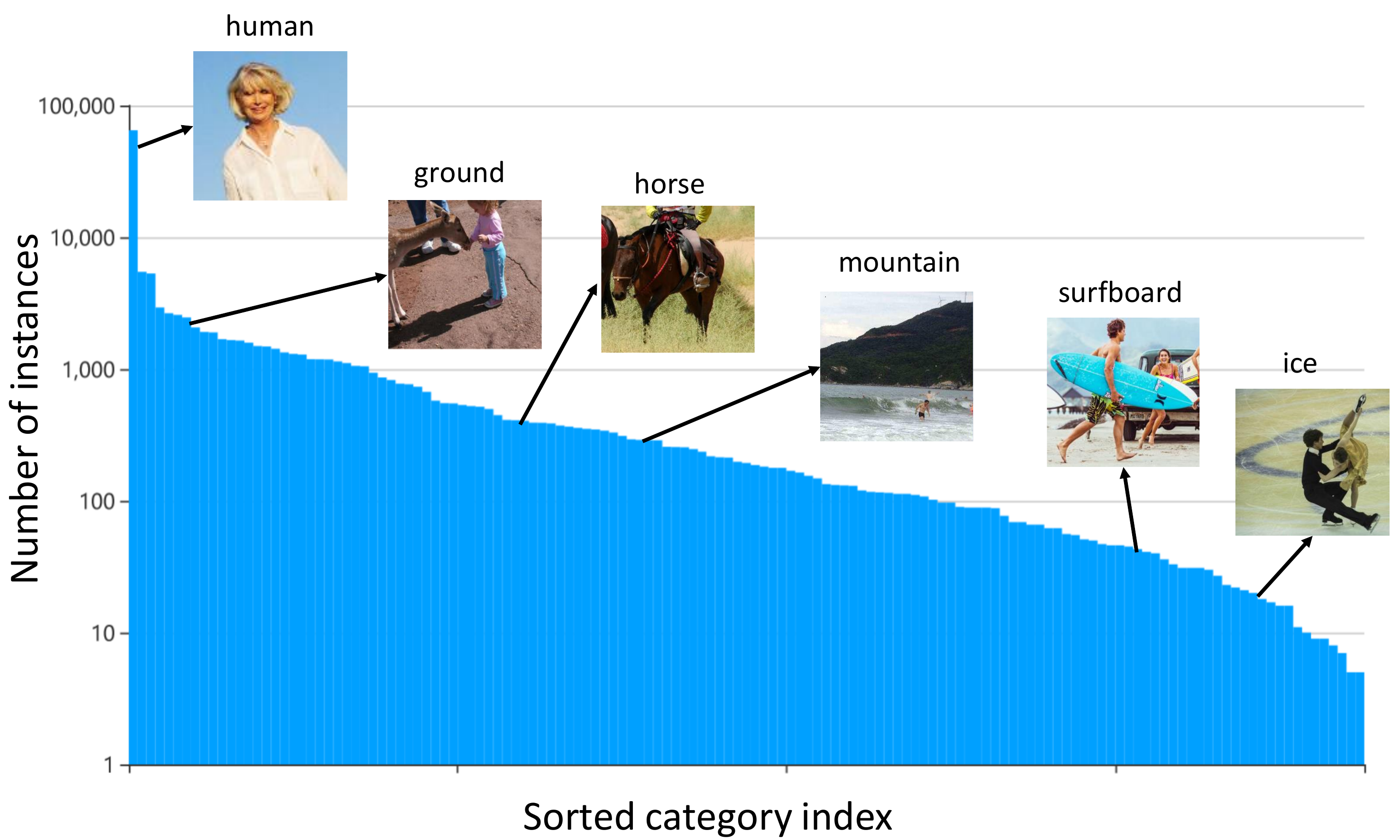}
	\caption{Distribution of things and stuff.}
	\label{fig:object_stuff}
\end{figure}

The  distribution of entities is shown in Figure \ref{fig:object_stuff}. The most frequent is ``human''. The ``ground'' also occurs frequently. On the contrary, ``surfboard'' and ``ice'' rarely occur.
The distribution of relations is shown in Figure \ref{fig:predicate} and \ref{fig:all_triplet}. The geometric relations are more frequent, and both geometric and action relations have long-tailed distributions.

\begin{figure*}[h]
	\centering
	\includegraphics[width=0.9\textwidth]{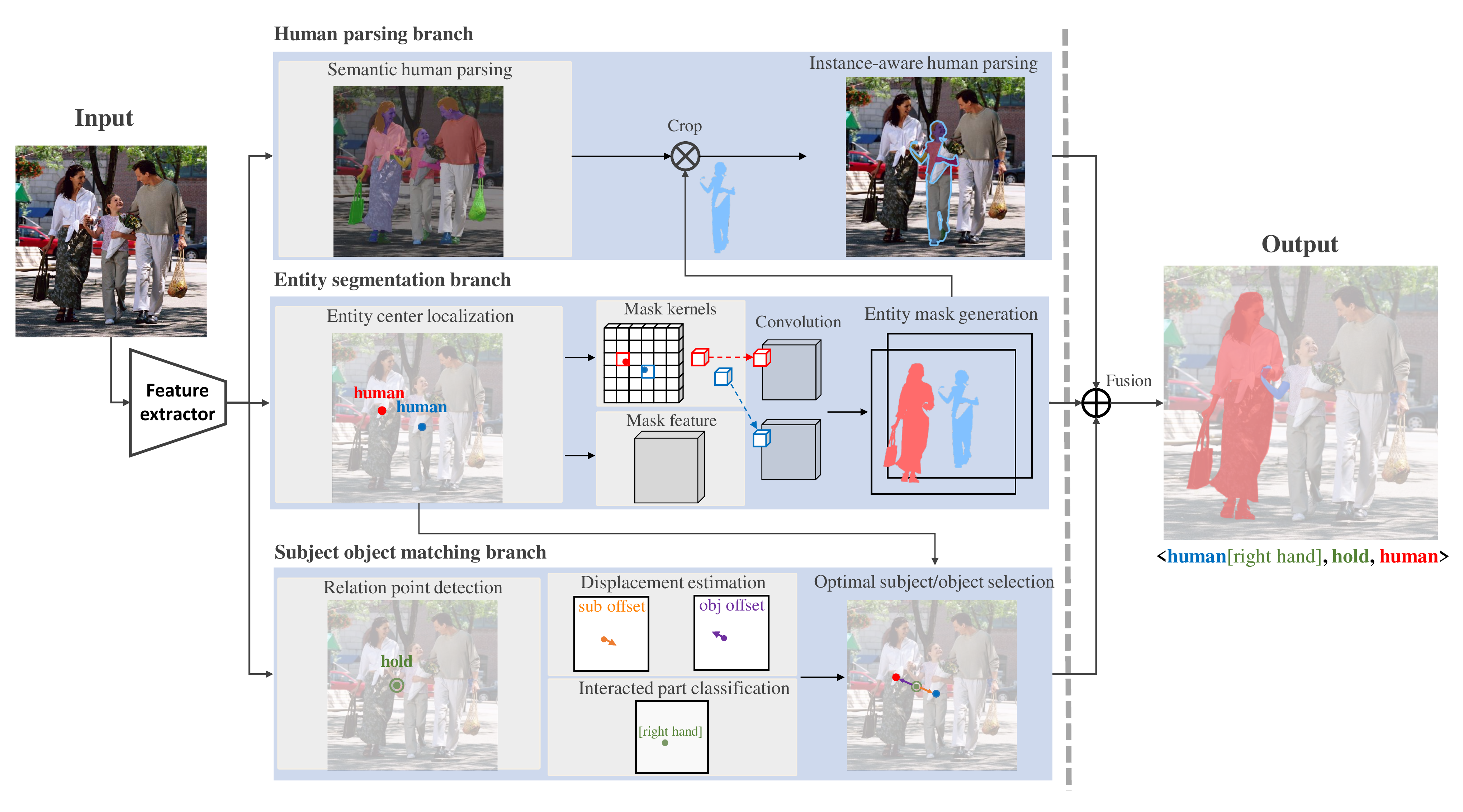}
	\caption{Framework of the proposed SMS. It is composed of a feature extractor and three parallel branches. The entity segmentation branch detects entities centers and describes them by masks. The subject object  matching branch pairs the corresponding subjects and objects and identifies their relations. The human parsing branch  segments the semantic human parts. All these intermediate results are fused to generate the HRS results.}
	\label{fig:framework}
\end{figure*}

PIC is split into $12,339$, $1,916$ and $2,867$ images for train, val and test set respectively.
Some examples of our extensively annotated PIC dataset are shown in Figure \ref{fig:final-dataset}.

\subsection{Comparisons with Existing Datasets}

As shown in Table \ref{tb:dataset_stat}, we compare our PIC dataset with several other popular datasets.
Similar with VRD datasets like Visual Genome \cite{krishna2017visual} and VRD \cite{lu2016visual}, PIC labels the relations among things and stuff.
Comparatively,  the HOI-det datasets like VCOCO \cite{gupta2015visual} and HICO-DET  \cite{chao2018learning} only label the relations between human and things.
In addition,  PIC has per-pixel masks to  describe the entities more precisely, especially when  the subjects or objects belong to \textit{stuff} type. V-COCO and Visual Genome also contain mask annotations brought from the original COCO dataset, but mask annotations in PIC have  higher quality and better boundary consistency.
PIC also labels  human poses and human parsing to facilitate the detailed inference of interacted human parts.
Moreover, PIC is a  high-resolution dataset with its average resolution significantly larger than other existing datasets, supporting fine-grained recognition of human and its context.

\section{Our method} \label{sec:task}
\subsection{Overview}

We  propose a single-stage Simultaneous Matching  and Segmentation (SMS) method to solve the HRS task.
The framework is shown in Figure \ref{fig:framework}. It contains a feature extractor  and three parallel branches including the entity segmentation branch, the subject object matching branch and the human parsing branch.
Given an input image $I \in \mathbb{R}^{H^{0} \times W^{0} \times 3}$, the extracted feature is $F \in \mathbb{R}^{H \times W \times d}$, where $\left( {H,W} \right) = \left( {H^{0}/s},{W^{0}/s} \right)$ and $s$ is the output stride. If not specified, the coordinates in the remaining of this paper are relative to the $H \times W$ feature size.

The entity segmentation branch detects the entities in the image and describes them with the pixel-level masks.
The subject object matching branch detects the existence of any relations and locates the corresponding subjects and objects.
For geometric relations, the subject are human instances, while for action relations, the subjects  are further defined as interacted human parts.
The human parsing branch decomposes humans into several  semantic human  parts, serving for interacted parts localization regarding the action relations.
Afterwards, the intermediate  results are fused to produce the HRS results. Next, we will describe our framework in detail.

\subsection{ Entity Segmentation Branch}
To output the  pixel-wise masks for the entities (subjects and objects),  SMS contains an entity segmentation branch with an entity center localization module and an entity mask generation module.

\textbf{Entity center localization:} We detect  the center coordinates $\left( x,y \right) \in \mathbb{R}^{2}$ and its class label $c$ for each entity.
Since directly detecting the center point is difficult, we follow the key-point estimation methods  \cite{newell2016stacked} to splat a point into a heatmap with a Gaussian kernel.
As shown in Figure \ref{fig:framework}, given an input image $I$,  the entity center heatmap ${Y} \in \mathbb{R}^{H \times W \times C_{ent}}$ is estimated, where $C_{ent}$ is the number of entity categories including  ``human'' and other $C_{ent}-1$  entity categories.
If the center of an entity belonging to the $c^{th}$ category is located at $\left( x^{0},y^{0} \right)$ in the input image $I$, then  the entity center heatmap ${Y}$   will have a high response value at $\left( {x,y,c} \right)$, where $\left( {x,y} \right) = \left( \left\lfloor {x^{0}/s} \right\rfloor,\left\lfloor {y^{0}/s} \right\rfloor \right)$.

To balance the positive and negative points, we  use focal loss \cite{law2018cornernet,zhou2019objects} on the center heatmap. Given the ground truth heatmap $\hat{Y}$ and the predicted heatmap $Y$, the mask center localization loss $L_{ent}$ is defined as

\begin{equation}
{
	L_{ent} = -\frac{1}{N^{pos}}\sum\limits_{xyc}
	\begin{cases}
		\left( 1 - {Y}_{xyc}\right)^{\alpha}{\log\left( {Y}_{xyc} \right)} \; &\text{if}~\hat{Y}_{xyc} = 1,
		\\[3mm]
		 \begin{gathered}
			({1 - \hat{Y}_{xyc}} )^{\beta}\left( Y_{xyc}\right)^{\alpha}\\
			{\log\left( {1 - {Y}_{xyc}} \right) }
		\end{gathered}&\text{otherwise}.
	\end{cases}
}
\label{equ:centernet}
\end{equation}\vspace{0.7ex}

In Equation (\ref{equ:centernet}),  $N^{pos}$ is the number of ground truth entities. The hyper-parameters $\alpha$ and $\beta$ are   set to $2$ and $4$.

\textbf{Entity mask generation:}  Inspired by SOLOv2 \cite{wang2020solov2}, we adopt a location-aware entity mask generation method that firstly learns mask kernels and mask features in parallel, and then applies $1 \times 1$ location-aware convolution  kernel on the mask features to produce the mask. More concretely,  the  {mask kernels} are of the size  $M_{kernel} \in \mathbb{R}^{S \times S \times D}$, where the input image is divided into $S \times S$ grids and each grid cell stands for a unique location-aware convolution kernel.
If the mask center of an entity falls into one grid cell at $(i,j)$, its corresponding location-aware mask kernel ${M^{i,j}_{kernel}} \in \mathbb{R}^{1 \times 1 \times D }$ is selected from  $M_{kernel}$. Then  ${M^{i,j}_{kernel}}$ is convolved with the mask feature $M_{feat} \in \mathbb{R}^{H \times W \times D}$ to generate the entity mask $m_{i,j}\in \mathbb{R}^{H \times W }$.

The loss function of entity mask generation is
\begin{equation}
L_{mask} = \frac{1}{N^{pos}}{\sum\limits_{(i,j)\in S^{pos}}{d_{mask}\left( m_{i,j},\hat{m}_{i,j} \right)}},
\end{equation}
where $N^{pos}$ is the number of ground truth entities and $S^{pos}$ is the set of center coordinates of ground truth entities.  $m_{i,j}$ and $\hat{m}_{i,j}$ are the predicted mask and ground truth mask. $d_{mask}\left( \cdot \right)$ can be any type of segmentation loss function. Here Dice Loss  is chosen.

\subsection{Subject Object Matching Branch} \label{sec:subject object Matching}
Suppose  $M$ entities are segmented. There exist $M(M-1)$ kinds of possible combinations. Traditional VRD methods  traverse all combinations, which is quite time consuming.

In order to pair the related subject and object, we introduce a relation point $\left( {x_{r},y_{r}} \right) = \left( \left\lfloor \frac{x_{s} + x_{o}}{2} \right\rfloor,\left\lfloor \frac{y_{s} + y_{o}}{2} \right\rfloor \right)$,
which is the midpoint of the corresponding subject center $\left( {x_{s},y_{s}} \right)$ and object center $\left( {x_{o},y_{o}} \right)$. Moreover, we estimate the displacement between the
relation point $\left( {x_{r},y_{r}} \right)$ and subject center $\left( {x_{s},y_{s}} \right)$, denoted as $\left( {d_{r}^{sx},d_{r}^{sy}} \right)$.
Thus the matched subject is supposed to be around $\left( {x_{r} + d_{r}^{sx},y_{r} + d_{r}^{sy}} \right)$.
 Similarity, we can estimate the displacement between the relation point  and object center, denoted as $\left( {d_r^{ox},d_r^{oy}} \right)$. Thus the matched object is supposed to be around $\left( {x_{r} + d_{r}^{ox},y_{r} + d_{r}^{oy}} \right)$.
Thus the relation point serves as the \emph{route} to match the corresponding subject $\left( {x_{r} + d_{r}^{sx},y_{r} + d_{r}^{sy}} \right)$  and object $\left( {x_{r} + d_{r}^{ox},y_{r} + d_{r}^{oy}} \right)$.
Next, we will elaborate how to detect the relation point and estimate the two displacements.

\textbf{Relation point detection:} we define the relation point as the middle of subject and object centers. Thanks to the large  respective field of the network, the relation point is able to  capture sufficient relation detection information.
The relation point is detected in a similar way with the above mentioned entity center localization. As shown in Figure~\ref{fig:framework}, a relation point heatmap $P \in \mathbb{R}^{H \times W \times C_{rel}}$ is generated, where $C_{rel}$ is the number of relation categories including action relations like ``hold'' and geometric relations like ``in front of''.
The responses around $\left( x_{r}, y_{r}, c \right)$ in $P$ will be high, where $c$ is the corresponding relation category.

\textbf{Displacements estimation:} In addition, for each relation $r$, we estimate the displacement from the relation point to the corresponding subject/object center, which is denoted as $\left( {d_{r}^{sx},d_{r}^{sy}} \right)$ and $\left( {d_{r}^{ox},d_{r}^{oy}} \right)$ respectively. Specifically, the model outputs two displacement offset maps $D_{s}$ and $D_{o} \in \mathbb{R}^{H \times W \times 2}$.

The relation point detection loss $L_{rel}$ is similar with $L_{ent}$ while the subject/object displacement estimation loss $L_{disp}$ is defined as
\begin{equation}
\begin{aligned}
	L_{disp} =  \frac{1}{N^{r}} \! \sum\limits_{r \in S^{rel}} (
		& | {d_{r}^{sx} - {\hat{d}}_{r}^{sx}} | + | {d_{r}^{sy} - {\hat{d}}_{r}^{sy}} |  \\[-3mm]
	  &+ | {d_{r}^{ox} - {\hat{d}}_{r}^{ox}} | + | {d_{r}^{oy} - {\hat{d}}_{r}^{oy}} | ) ,
\end{aligned}
\end{equation}
where $N^{r}$ is the number of ground truth relations and $S^{rel}$ is the set of ground truth relations.

\textbf{Interacted part classification:} When the relation is an action relation, the subject of the triplet is further refined as the human parts  making the action.
For example,  $\langle$human, hold, cup$\rangle$ can be further refined as $\langle$human [left arm], hold, cup$\rangle$ and  $\langle$human, stand on, grass$\rangle$ can be further refined to $\langle$human [right shoe], stand on, grass$\rangle$. To this end, we output the probability of each human part  participating in the action. It is treated as a multi-label classification task. For every relation point $\left( {x_{r},y_{r}} \right)$,  the branch produces a multi-hot classification result $s_{r} \in \mathbb{R}^{C_{part}}$, where $C_{part}$ is the number of human part categories.

The loss function of interacted part classification is defined as
\begin{equation}
L_{part} = \frac{1}{N^{r}}{\sum\limits_{r \in S^{rel}}{BCEloss\left( s_{r}, \hat{s}_{r} \right)}},
\label{equ:bceloss}
\end{equation}
where $N^{r}$ is the number of relation categories and $ \hat{s}_{r}$ is the groundtruth part classification label.  Note that we use BCEloss loss here because multiple parts may relate to the same object. For instance, both left arm and right arm may hold the cup.

\subsection{Human Parsing Branch}

To generate the segmentation mask for each human part, e.g. left leg, right arm, we generate  human parsing results $P_{sem} \in \mathbb{R}^{H \times W \times C_{part}}$,  where $C_{part}$ is the number of human part categories. The human parsing loss function $L_{human}$ is defined as 
\begin{equation}
	L_{human} = L_{ce} + L_{iou},
\end{equation}
where $L_{ce}$ is the cross entropy loss widely used in semantic segmentation and $L_{iou}$ is the Lovasz loss  \cite{berman2018lovasz} directly optimizing the intersection over union.
Then, the subject segmentation result is used to crop on the human parsing result $P_{sem}$   to get the instance-specific human part segments.

\subsection{Loss and Inference}
The total loss of our SMS method is the weighted sum of the above mentioned losses:
\begin{equation}
L\! = \!L_{ent} + L_{rel} + \lambda_{1}L_{mask} + \lambda_{2}L_{disp}  + \lambda_{3}L_{part} + \lambda_{4}L_{human},
\end{equation}
where $\lambda_{1}$, $\lambda_{2}$, $\lambda_{3}$ and $\lambda_{4}$  are weights to balance the loss.

During inference,  firstly we pick top$K_s$, top$K_o$ and top$K_r$ local maximum activation for subject center, object center and relation point respectively, where $K_s$, $K_o$ and $K_r$ are hyperparameters and set empirically. Each detected relation point finds its subject and object through an optimal subject/object selection process  by Equation (\ref{equ:matching}). During the process, the estimated subject center of a relation $r$ is obtained by summing the entity center localization results and displacements estimation results $\left( x_{r}\! +\! d_{r}^{sx},y_{r}\! + \!d_{r}^{sy} \right)$. Then the optimal subject $	\left( {x_{r}^{s},y_{r}^{s}} \right) $ is selected from all the detected subjects  $S^{sub}$ according to the distances from their centers to the  estimated subject center.
 \begin{equation}
	\left( {x_{r}^{s},y_{r}^{s}} \right) = \! \underset{(x,y,c) \in S^{\mathit{sub}}}
	{argmin}\frac{1}{Y_{x,y,c}}\!
	\left| \left( x,y \right)\! -\! \left( x_{r}\! +\! d_{r}^{sx},y_{r}\! + \!d_{r}^{sy} \right)\right|,
	\label{equ:matching}
\end{equation}
where $S^{sub}$ is the set of detected subject center points and ${Y_{x,y,c}}$ is the value at $(x,y,z)$ in the entity center heatmap $Y$. The optimal object $	\left( {x_{r}^{o},y_{r}^{o}} \right) $  is selected in the same way. After localizing the subject and object,  their mask kernels are selected according to the grid $(i,j)\in {S \times S }$ which $\left( {x_{r}^{s},y_{r}^{s}} \right)$  and $\left( {x_{r}^{o},y_{r}^{o}} \right)$ fall in. Then the subject mask $m_{sub}$ and object mask $m_{obj}$  are produced by applying  the dynamically selected  mask kernels on the mask feature.

For action relation triplets, the results are in the form of  $\langle$human [part], action, entity$\rangle$.  Specifically, we  get human part classification results  $s_{r}$ to identify the interacted part. Then, the interacted human part segments are generated by fusing the subject masks $m_{sub}$, semantic human parsing result $P_{sem}$ and part classification results $s_r$.

\section{Experiments} \label{sec:Experiments}
We first compare the proposed SMS with baselines in HRS task over the  PIC dataset.
To  further show its generalization  ability, we also evaluate SMS in the Relation Segmentation (RS) task on both PIC and V-COCO datasets \cite{gupta2015visual}.

\begin{table*}[htb!]
	\renewcommand\arraystretch{1.3}
	\vspace{8mm}
	\begin{center}
		\small
		\begin{tabular}{cc|cccc|cccc|c}
			\hline
			&Relation &\multicolumn{4}{c|}{RS(\%)} & \multicolumn{4}{c|}{HRS(\%)}                                               &                                                \\
			Method           		& Backbone& mR@$25$      & mR@$50$      & mR@$100$      & AR & \multicolumn{1}{c}{mR@$25$} & \multicolumn{1}{c}{mR@$50$} & \multicolumn{1}{c}{mR@$100$}  & \multicolumn{1}{c|}{AR} & Running Time\\ \hline
			$m$-MOTIFS\cite{zellers2018neural}                   &VGG-16& 19.89         & 27.90         & 32.78          & 26.86 & 12.62                        & 18.01                        & 21.43                         & 17.36  & 107 ms+\emph{163ms}                                                \\
    			$m$-MOTIFS*              &ResNet-50& 21.36         & 28.57         & 33.54          & 27.82 & 13.90                        & 19.10                        & 22.67                         & 18.56  & 109 ms+\emph{163ms}                                              \\
    			$m$-KERN\cite{chen2019knowledge}			        &VGG-16& 21.27         & 29.30         & 34.55          & 28.37 & 13.98                        & 19.41                        & 22.97                        &   18.79  & 149 ms+\emph{163ms}                                                 \\
    			$m$-KERN*		        &ResNet-50& 22.32         & 29.65         & 34.76          &  28.91 & 14.14                        & 19.37                        & 23.31                        & 18.97  & 159 ms+\emph{163ms}                                               \\
			$m$-PMFNet\cite{wan2019pose}                  &ResNet-50& 23.01         & 27.85         & 29.50          & 26.79 & 14.90                        & 17.74                        & 18.87                         & 17.17  & 63 ms+\emph{163ms}                                              \\ \hline
			SMS                    &ResNet-50& 24.42         & {30.52}                  & 34.96 & 29.97                        & 15.81                        & 19.42                         & 22.45  &19.23  & \textbf{{28 ms}}                                   \\
			SMS                   &DLA-34& {25.25}         & 30.01                   & {34.97} & {30.08}                        & {16.35}                        & {19.45}                         & {23.38}  &{19.73}  & 31 ms                                  \\
		SMS-L &ResNet-50 &\textbf{26.58} &\textbf{32.8} &\textbf{37.21} &\textbf{32.21} &\textbf{17.63} &\textbf{21.53} &\textbf{24.90}&\textbf{21.35} & 81 ms \\ 
			\hline
		\end{tabular}
	\end{center}
	\vspace{-2mm}
	\caption{Comparisons on PIC test set. }
\label{tb:predictions}
\end{table*}

\begin{table}[htb!]
	\renewcommand\arraystretch{1.3}
	\vspace{2mm}
	\begin{center}

		\small

		\begin{tabular}{p{3.0cm}<{\centering}|p{2.5cm}<{\centering}}
			\hline
			Method           		& $AP_{role}$(\%)     \\ \hline
			$m$-MOTIFS\cite{zellers2018neural}                   & 40.74          \\
			$m$-KERN\cite{chen2019knowledge}			        & 41.31       \\
			$m$-PMFNet\cite{wan2019pose}                  & 43.13    \\ \hline
			SMS                    & \textbf{44.13}     \\
			\hline
		\end{tabular}
	\end{center}
	\vspace{-2mm}
	\caption{Comparison on VCOCO test set.  }
\end{table}
\label{tb:vcoco_predictions}

\subsection{Implementation Details}
We use   DLA-34  network \cite{yu2018deep} and ResNet-50  \cite{he2016deep} as our backbones  to show the generalization of the proposed SMS framework.
We modify the light-weight DLA network  to contain a progressive refinement architecture following CenterNet~\cite{zhou2019objects}. We use ResNet with FPN  architecture \cite{lin2017feature} to enable multi-scale predictions. The FPN layers are fused to generate single scale features according to \cite{kirillov2019panoptic}.  The parameters of the backbone are initialized from the pretrained weights on ImageNet dataset~\cite{deng2009imagenet}.
All the parallel branches except the mask generation branch share the same architecture, which is composed of two (three in the relation point detection branch) $3\times3$  convolutional layers followed by RELU and one $1\times1$ convolutional head for prediction. In the entity segmentation branch, the heads of mask kernels and mask feature are both composed of two $3\times3$ and one $1\times1$ convolutional layer followed by group normalization and RELU, where the first convolution layer is CoordConv~\cite{wang2020solov2}. If not specified, these branches are initialized randomly and trained from scratch.

By widening and deepening  three parallel branches  of  SMS, we obtain a larger model denoted as SMS-L.  SMS-L shares the same feature extraction backbone as SMS. The difference is that SMS-L contains more parameters and thus introduces more computational cost to each parallel branch. \textcolor{black}{More spcifically, for SMS-L, we stack 5 more convolutional layers followed by group normalization and ReLU before entity segmentation branch, subject object matching branch and human parsing branch respectively, with progressive summation operation for feature fusion. The numbers of convolutional layers in entity segmentation branch, subject object matching branch and human parsing branch are set to 3, 3 and 5 respectively. We also expand the feature channel in the three parallel heads from 128 to 384}.

For the PIC dataset, we train for $140$ epochs  with Adam~\cite{kingma2014adam}. The initial learning rate is 3e-4, and batch size is $60$. For the V-COCO dataset, we also train  for $140$ epochs. The initial learning rate is 4e-4, and batch size is  $40$.  For both datasets, the learning rate is divided by $10$ at $90^{th}$ and $110^{th}$ epoch respectively. The loss  weights $\lambda_{1}$, $\lambda_{2}$, $\lambda_{3}$ and $\lambda_{4}$  are set to $3$, $1$, $20$ and $5$  respectively in  experiments.

The speed of our SMS model is evaluated on a single RTX $2080$Ti GPU, with Pytorch $1.1$ and CUDA $10.2$. When evaluating the inference speed,  we only calculate the  running time of the models and ignore the time consumption of the result saving process.

\subsection{Baselines }
 We design several baselines for both RS and  HRS tasks.

For the  RS task, we  modify  the  state-of-the-art VRD methods MOTIFS\cite{zellers2018neural}, KERN\cite{chen2019knowledge} and HOI-det  method PMFNet\cite{wan2019pose}  to produce the entity segmentation results.
These methods rely on a strong object detector Faster R-CNN \cite{ren2015faster} to generate object candidates and then infer their relations, which follow a two-stage pipeline.
These modified RS baselines are denoted as $m$-VRD, $m$-MOTIFS and $m$-PMFNet respectively.
More specifically, we  replace the Faster R-CNN in these methods with Mask R-CNN~\cite{he2017mask} with ResNet-50 backbone to generate entity masks. The $m$-PMFNet does not use human pose features for fair comparison.

For the HRS task, to produce human part level segmentation, a human parsing head is attached to the Mask R-CNN to output instance-level human parsing results. To identify which is the interacted part, several additional fully connected layers are added in parallel with the original relation prediction head. The human parsing results, interacted part classification results,  together with the relation prediction results are merged to form the final HRS results.

\subsection{Evaluation Metric}
The widely used average precision metric is not suitable for PIC on account of some potential relations not annotated which may cause a lot of misclassified ``false positive'' samples.
We use mean $\text{Recall}@K$ (abbr.  mR) rather than $\text{Recall}@K$ to evaluate both RS and HRS following \cite{tang2020unbiased}, considering the reporting bias as discussed in \cite{misra2016seeing}.

For RS and HRS, a predicted relation triplet is regarded as a True Positive (TP) sample if it matches with any ground truth relation triplet in following respects: 1) the subject, object and relation categories are both consistent with the ground truths, and 2) the IoU between predicted truth subject mask and ground subject mask is above a threshold $\tau_{iou}$, and the same case for the object.

Mean Recall@$K$ is the mean of Recall@$K$ per relation category. The most $K$ confident relation triplets are used for evaluation.
We calculate mR@$K$ as
\begin{equation}
\normalsize
\text{mR}@K\! =\! \frac{1}{\left| S^{\tau} \right|} \sum\limits_{\tau_{iou} \!\in S^{\tau}} \frac{1}{C}
\sum\limits_{c \in [1...C]}
\Big( {\sum\limits_{p \in P_{c}}\mathbbm{1}_{\{p ~\text{is TP}\}}} \Big) / \left| G_{c} \right|,
\end{equation}
where $\left| \cdot \right|$ is the cardinality of a set, $S^{\tau}\!=\!\{ 0.25,0.5,0.75\}$ and $\tau_{iou}$ is the IoU threshold for assigning predicted subjects and objects to ground truth, $C$ is the number of relation categories, $P_{c}$ is the set of predicted triplets belonging to relation type $c$ and $\sum\limits_{c} \left| P_{c} \right| \leq K$, $G_{c}$ is the set of ground truth triplets belonging to the relation type $c$.

The corresponding response values in the heatmap are the confidences of detected entities and relationships. Confidence of a relation triplet is calculated by multiplying the confidences of subject, object and relationship.
To comprehensively evaluate the method,  we also adopt AR which is  the mean of mR@$25$, mR@$50$ and mR@$100$.

\subsection{Experiments on PIC Dataset}
We evaluate performance on both RS and HRS tasks.
Different from VRD, RS  uses pixel-level masks to represent entities rather than bounding boxes.
The quantitative results are shown  in Table \ref{tb:predictions}.
In their original papers, MOTIFS \cite{zellers2018neural} and KERN \cite{chen2019knowledge}  use VGG-16 as the feature extractor backbone.
For fair comparison, we re-implement these two baselines using more powerful ResNet-50, which are denoted as $m$-MOTIFS* and $m$-KERN*.
For the RS task,  SMS-L achieves the best performance of $32.21$\% AR in RS, which is $3.3$\% higher than $m$-KERN* and $4.39$\% higher than $m$-MOTIFS*. Moreover, the improvements over baselines are consistent in all evaluation metrics from  $\text{mR}@25$ to $\text{mR}@100$.
For the  HRS task,  SMS-L  also achieves the best accuracy of  $21.35$\% AR, which is $2.38$\% higher than $m$-KERN* and  $2.79$\% higher than $m$-MOTIFS*.
It indicates that SMS-L is able to recognize humans, objects and their relations at a finer level.
Moreover, SMS-L outperforms  SMS  at the cost of inference latency.

We have implemented SMS with DLA-34 and ResNet-50 to show its generalization ability. 
For $m$-MOTIFS* and $m$-KERN*, the results with ResNet-50 are better than VGG-16.
For SMS, ResNet-50 can produce faster inference but slightly lower performance than DLA-34.
With both backbones, SMS achieve higher AR  than all baselines in both RS and HRS settings.
It shows that SMS is a general framework and not limited to a specific backbone.

Moreover, we   evaluate  the inference speed using  the $\text{mR}@25$ setting  in HRS.
Specifically, the hyper parameters $K_{s}$,  $K_{o}$ and $K_{r}$ are set to $25$, $10$ and $30$ respectively.
The running time does not include the time consumption of data reading and result saving.  
Note that the running time of baselines include two parts. For example, the running time of $m$-PMFNet is composed of  relation prediction time (63ms) and  instance segmentation time (163ms). 
SMS reaches $32$ FPS (with DLA-34) and $36$ FPS (with ResNet-50)  and is the only method achieving  \emph{real-time} inference speed.  ResNet-50 is slightly faster than DLA-34, but the AR is slightly lower. The results show that the single-stage architecture of SMS achieves better speed-accuracy trade off than the two-stage baselines.

Table \ref{tb:geoandact} shows the results of geometric and action relations. 
The performance of SMS-L exceeds the baselines significantly in terms of both geometric and action relations. 
Especially for action relations, our method performs much better than the  baselines. The reason  lies in that we predict relations directly from the whole feature map, reducing the error introduced by imperfect detection results and retaining more context beneficial for action relation prediction. In sum, our method performs the best for both geometric and action relations, in  both RS and HRS tasks with much less inference time.

\begin{table}[h]
	\renewcommand\arraystretch{1.3}.
\caption{Performance on  geometric (Geo) and action (Act) relations with ResNet-50 backbone.}
	\begin{center}
		\small
		\resizebox{0.488\textwidth}{!}{
			\begin{tabular}{c|ccc|ccc}
				\hline
				&\multicolumn{3}{c|}{RS} & \multicolumn{3}{c}{HRS} \\
				Method& Geo& Act& Mean&  Geo& Act& Mean \\ \hline
				$m$-MOTIFS*\cite{zellers2018neural}                        &32.84 &24.35 &28.60 &32.64 &8.67 &20.75
				\\
				$m$-KERN*\cite{chen2019knowledge}                          &32.68 &26.17 &29.43 &32.68 &8.78 &20.73				
				\\
				$m$-PMFNet\cite{wan2019pose}      &26.89 &26.72 &26.81 &26.89 &10.43 &18.66
				\\ \hline
				{SMS-L}                            &\textbf{33.68} &\textbf{31.19} &\textbf{32.44} &\textbf{33.68} &\textbf{12.82} &\textbf{23.25}
				\\
				\hline          
			\end{tabular}
		}
	\end{center}
	\vspace{0mm}
	\label{tb:geoandact}
\end{table}

We define the categories of relationships which appear less than $800$ times in our dataset as the rare relations. Table \ref{tb:rare_2} shows the comparison on rare and non-rare relations. It can be seen that our larger model SMS-L  achieves best performance on both  rare relations and non-rare relations.

\begin{table}[h]
	\renewcommand\arraystretch{1.3}
\caption{Performance on  rare and non-rare relations with ResNet-50 backbone.}
	\begin{center}
		\small
		\resizebox{0.488\textwidth}{!}{
			\begin{tabular}{c|ccc|ccc}
				\hline
				&\multicolumn{3}{c|}{RS} & \multicolumn{3}{c}{HRS}\\
				Method& Rare& Non-Rare& Mean&  Rare& Non-Rare& Mean\\ \hline
				$m$-MOTIFS*\cite{zellers2018neural}&
				17.30&35.11&26.20&7.56&26.17&16.87\\
				$m$-KERN*\cite{chen2019knowledge}&
				19.44&35.12&27.28&7.43&25.86&16.65\\
				$m$-PMFNet\cite{wan2019pose}&
				21.05&31.07&26.06&11.62&21.00&16.31\\ \hline
				{SMS-L}& 
				\textbf{22.61}&\textbf{38.86}&\textbf{30.73} &\textbf{13.96}&\textbf{26.47}&\textbf{20.22}\\	
				\hline          
			\end{tabular}
		}
	\end{center}
	\label{tb:rare_2}
\end{table}

\subsection{Experiments on V-COCO Dataset}
The large-scale HOI-det dataset V-COCO~\cite{gupta2015visual}  is annotated on a subset of COCO. V-COCO treats the objects associated with actions as different semantic roles including a direct object or an instrument object.
For V-COCO, evaluation is conducted on these two types of roles separately.
We extend V-COCO annotations with the original COCO annotations to generate RS annotations by simply replacing the bounding box annotations  with mask annotations.
Because V-COCO does not contain human parsing and interacted parts annotations, we do not evaluate the performance of HRS in this experiment. We modify the standard evaluation metric $AP_{role}$ proposed for V-COCO benchmark  to evaluate the RS results.
An estimated RS result is treated as true positive if both the predicted subject mask and object mask have IoU over  $0.5$ with any ground truth subject  mask  and the corresponding ground truth object mask given the action category.

We  evaluate SMS on the customized V-COCO dataset with the baselines.
Table \ref{tb:vcoco_predictions} reports the performances on the test set.
Slightly different from Table \ref{tb:predictions}, the performance of HOI-det method $m$-PMFNet is better than the VRD method $m$-KERN here.
The observation accords with the intuition since KERN is designed for VRD  and PMFNet  for HOI-det.
Table \ref{tb:vcoco_predictions} shows SMS is able to achieve $44.13\%$ $AP_{role}$  with a performance gain of $1.00$\% $AP_{role}$ over $m$-PMFNet and  $2.82\%$ over $m$-KERN.
The improvements show  the effectiveness and generalization ability of SMS.

\subsection{Ablation Studies}
Here we evaluate different components of SMS on PIC test set. According to Table~\ref{tb:predictions}, DLA-34 achieves higher performance than ResNet-50. Thus we use DLA-34 backbone for the ablation studies.
If not specified, all the results are evaluated  in terms of the AR metric.

\subsubsection{Feature Extractor}

We use ResNet \cite{he2016deep}  and  DLA network \cite{yu2018deep} as the feature extraction backbones.
Both backbones can generate multi-level features of different scales. We further explore how SMS works with the largest single level feature map or  multi-level feature fusion.
To fuse  multi-level features, we put the  multi-level features  through a series of   convolution layers and upsample layers, and then fuse them with  summation (for ResNet) or concatenation (for DLA network)~\cite{kirillov2019panoptic}.

As shown in Table \ref{tb:predictions}, SMS with ResNet  achieves   higher inference speed while SMS with DLA  achieves  higher accuracy.
The DLA is slightly slower than ResNet. This is because  to enlarge the receptive field for the light-weight DLA network, we induce Deformale ConvNets~\cite{zhu2019deformable} in the DLA architecture following~\cite{zhou2019objects}.
Moreover,  SMS with both backbones outperforms baselines with real-time inference speed, revealing the generalization of the proposed framework.
From Table \ref{tb:abl_feature_extractor}, we can see multi-level features fusion  also brings some performance gains for both backbones in both tasks.

\begin{table}[htb!]
\renewcommand\arraystretch{1.3}
	\vspace{3mm}
	\begin{center}
		\small
		\begin{tabular}{cc|cc}
			Feature Extractors           		&& RS(\%)       & HRS(\%)        \\ \hline
			DLA-34  single-level                && 29.87         & 19.73        \\
			+multi-level feature fusion			        && \textbf{30.08}
         & \textbf{19.73}         \\ \hline
			ResNet-50   single-level          && 29.00 & 19.08 \\
			+ multi-level feature fusion  && 29.97 &   19.23  \\
			\hline
		\end{tabular}
	\end{center}
	\vspace{-2mm}
	\caption{Comparisons of different feature extractors. }
	\label{tb:abl_feature_extractor}
\end{table}

\subsubsection{Entity Segmentation Branch}
To validate the generalization ability of SMS, we  replace the entity segmentation branch with two  one-stage instance segmentation methods including  PolarMask \cite{xie2020polarmask} and SOLO~\cite{wang2019solo}.
PolarMask \cite{xie2020polarmask} tackles the mask generation problem as the  contour estimation problem and regresses contour polygons from the instance center.  SOLO  \cite{wang2019solo} learns  the spatial-sensitive instance features and predicts the pixel-level masks for each instance conditioned on their locations.
SMS adopts the SOLOv2	\cite{wang2020solov2} to segment the entity.
We only use the largest single-level feature map as the feature extractor.

Table \ref{tb:abl_mask_method}  demonstrates that  SMS can generalize to different mask generation methods.
In detail, SMS with PolarMask underperforms the other two models. It is mainly because PolarMask approximately represents the mask of the entity by polygons which have a fixed number of vertices (typically $36$). The approximate representations have a limited expressive ability especially when describing \textit{stuff} classes. Actually, it is in agreement with our claim that box-like representation  is not able to accurately describe the entity shape, which is exactly the motivation of our introducing the relation segmentation task.
Moreover, the experiment results indicate the SMS framework is very flexible and its performance  can be further boosted along with the latest developments of segmentation methods.

\begin{table}[htb!]
\renewcommand\arraystretch{1.3}
	\vspace{3mm}
	\begin{center}
		\small
		\begin{tabular}{cc|cc}
		Segmentation	Method           		&& RS(\%)       & HRS(\%)        \\ \hline
			PolarMask  \cite{xie2020polarmask}              && 19.88         & 13.75        \\
			SOLO   \cite{wang2019solo}                 && 27.66         & 18.11        \\
			SMS		        && \textbf{29.87}         & \textbf{19.73}
			        \\
			\hline
		\end{tabular}
	\end{center}
	\vspace{-2mm}
	\caption{Comparisons of different mask generation methods. }
	\label{tb:abl_mask_method}
\end{table}

\subsubsection{Subject Object Matching Branch}
We study the effectiveness of displacement estimation and interacted part classification in SMS.

\textbf{Displacement estimation:} As introduced in Section \ref{sec:subject object Matching}, we estimate a subject offset  $\left( {d_{r}^{sx},d_{r}^{sy}} \right)$ and an object offset  $\left( {d_{r}^{ox},d_{r}^{oy}} \right)$ for the relation point $\left( {x_{r},y_{r}} \right)$.
Another strategy is to estimate the  object offset with the negative of subject offset $\left( {d_{r}^{ox},d_{r}^{oy}} \right)=-\left( {d_{r}^{sx},d_{r}^{sy}} \right)$.
As shown in Table \ref{tb:abl_matching}, if estimating the subject offset only, the performance is nearly the same as estimating  both subject  and object offsets ($29.87\%$ vs $29.71\%$ in RS task and $19.74\%$ vs $19.65\%$ in HRS task).
It shows that SMS is able to accurately estimate the  relation point which is the midpoint of the related  subject and object pair. Therefore there is no obvious differences  between estimating two offsets and estimating subject  offset only.

\begin{figure*}[tp!]
	\centering
	\includegraphics[width=.7\textwidth,height=1.55in]{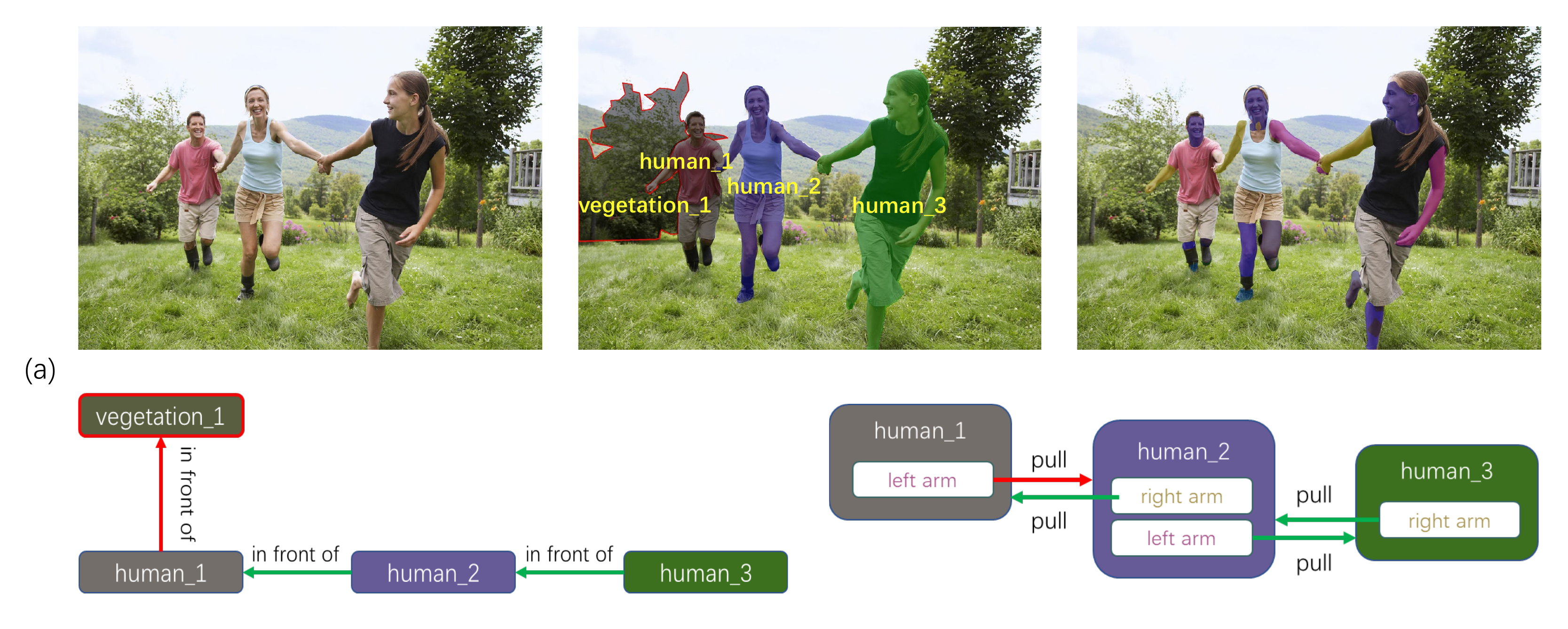}
	\includegraphics[width=.7\textwidth,height=1.55in]{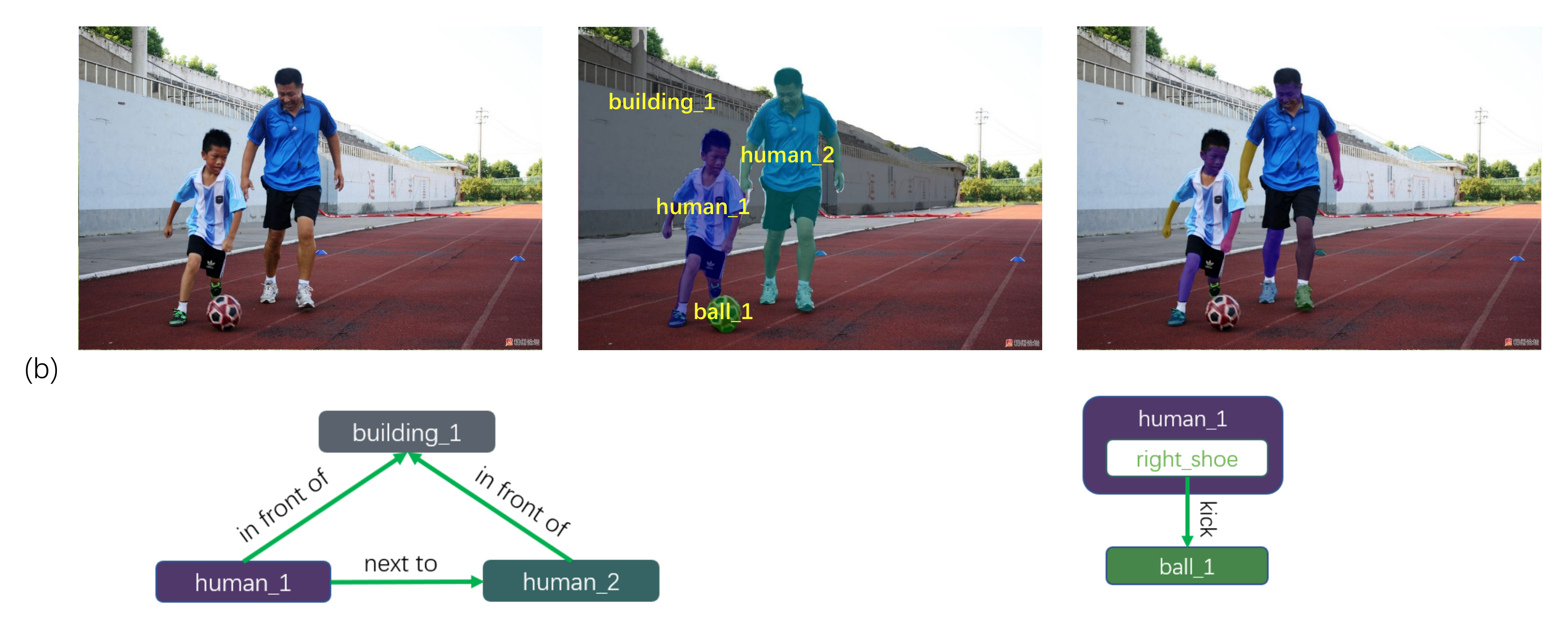}
	\includegraphics[width=.7\textwidth,height=1.75in]{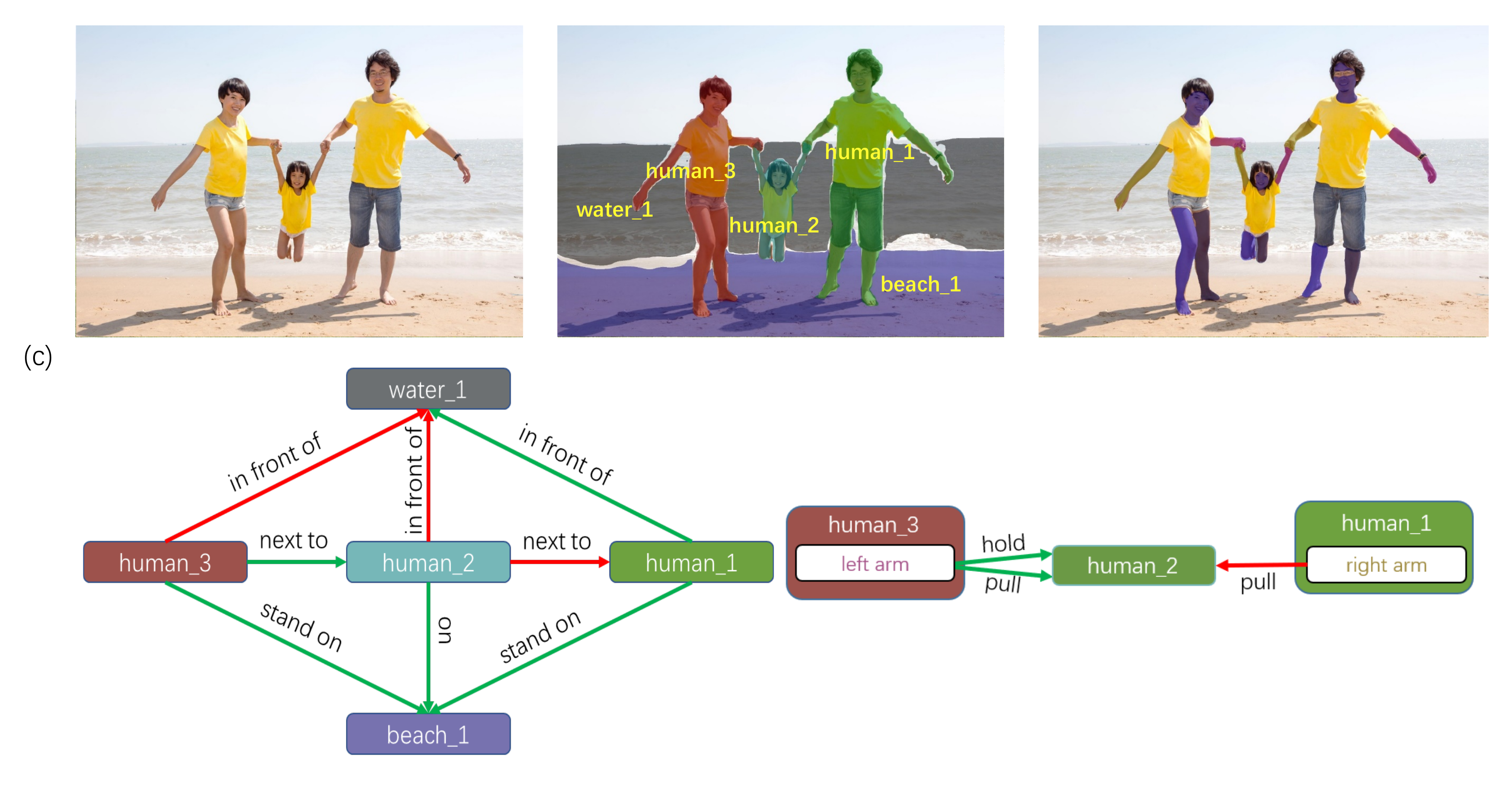}
	\includegraphics[width=.7\textwidth,height=1.85in]{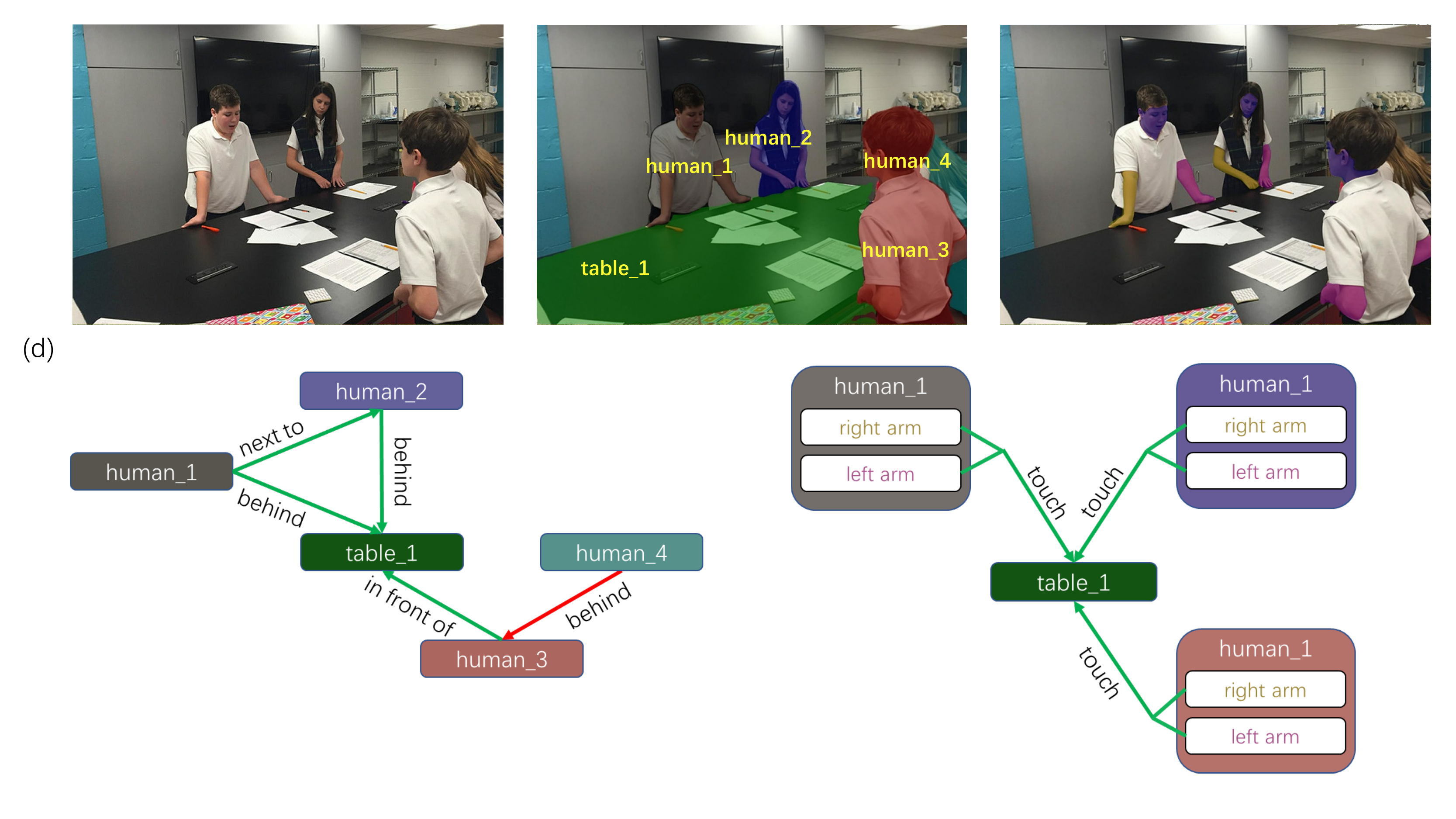}
	\includegraphics[width=.7\textwidth,height=1.85in]{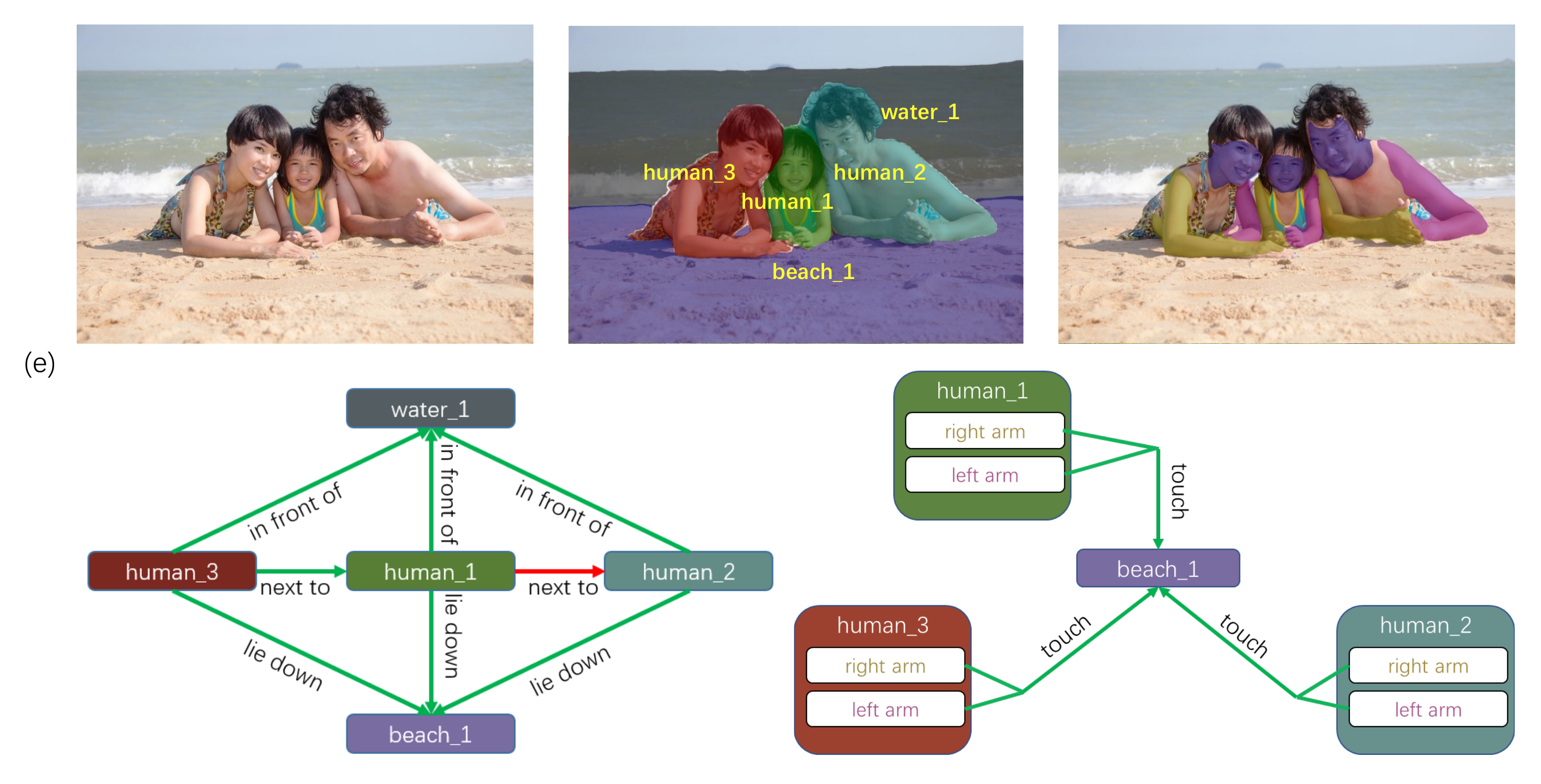}
	\caption{Visualization of  HRS results  on PIC test set.  For each image, only $25$ most confident relation triplets are shown. In each result set, the upper row  shows the original image,  entity segmentation masks and  interacted human part segments. The lower row are the  geometric  and action relation graph.
	In both graphs,  green and  red arrows  stand for true positives and false negatives.
	Nodes with red border are ground truth  masks not matched by the entities in the top 25 relations.  }
	\label{fig:result}
\end{figure*}

\begin{table}[htb!]
\renewcommand\arraystretch{1.3}
	\vspace{3mm}
	\begin{center}
		\small
		\begin{tabular}{cc|cc}
		Displacement Estimation	Method           		&& RS(\%)        & HRS(\%)        \\ \hline
			subject offset + object offset && \textbf{29.87}         & \textbf{19.73}        \\
			subject offset only                   && {29.71}         & {19.65}        \\
			\hline
		\end{tabular}
	\end{center}
	\vspace{-2mm}
	\caption{Comparisons of different displacement estimation strategies. }
	\label{tb:abl_matching}
\end{table}

\textbf{Interacted part classification:}
SMS also estimates the interacted part for each relation point.
We compare with three other straightforward statistic-based baselines including   ``Sample by distribution'', ``Most frequent'' and ``Nearest Part''.
In detail, we first calculate the frequencies/distribution of each interacted part category given the action relations. For example,  ``left arm''  co-exists more frequently with the action ``hold'' than ``left shoe''.
``Sample by distribution'' means sampling the interacted part category   according to their  distribution conditioned on the estimated action.
``Most frequent'' means directly selecting the most frequent interacted part category conditioned on the estimated action.
``Nearest Part'' determines the interaction body part according to their distance to the interaction point.

We evaluate the three interacted parts classification methods and the results are shown in Table \ref{tb:abl_part_classification}.
AR is evaluated only for  action relations and geometric relations are ignored.
The results show that  ``Sample by distribution'' and ``Most frequent'' are able to correctly predict the interacted human parts to some extent because of the data bias.
However, our SMS outperforms ``Sample by distribution'', ``Most frequent'' and ``Nearest Part'' by $2.64$\%,  $1.56$\% and $3.14$\% respectively.
It shows that the BCEloss in Equation \ref{equ:bceloss} applied on the relation point can guide our model to  distinguish interacted parts.
The average recall of ``Nearest Part'' is   $9.08\%$. This inferior performance may result from the diversity of the action appearance and the error of human parsing prediction.

\begin{table}[htb!]
	\renewcommand\arraystretch{1.3}
	\vspace{3mm}
	\begin{center}
		\small
		\begin{tabular}{c|c}
			\hline
		Classification	Method           		& HRS(\%)  \\ \hline
			Sample by distribution          	& 9.58  \\ 
						Most frequent                &  10.66     \\ 
				Nearest part        & 9.08                     \\
						SMS                   & \textbf{12.22}    \\
			\hline          
		\end{tabular}
	\end{center}
	\vspace{-2mm}
	\caption{Comparison of different  interacted part classification methods in HRS task. Evaluation is conducted on action relations only. }
	\label{tb:abl_part_classification}
\end{table}

\subsection{Qualitative Analysis}

Five HRS results on PIC dataset  are  shown in Figure \ref{fig:result}.
We choose the top $25$ highest score relations for each example image. 
Each result set contains five elements, including the original image,  entity segmentation masks,  interacted human part segments, the  geometric  and action relation graph.  The green and  red arrows in both relation graphs  stand for true positives and false negatives relations respectively.  The nodes with
non-red border represent entities among the predicted top 25 triplets which have IoU over 0.5 with ground truth masks, while the nodes with red border represent ground truth not matched with any entities in the top 25 predicted triplets.

In Figure~\ref{fig:result}a, SMS can correctly  estimate
fine-grained human-centric relations, such as $\langle$human\_2 [right arm], pull, human\_1$\rangle$ and $\langle$human\_2 [left arm], pull, human\_3$\rangle$. 
Notably, SMS can correctly estimate the relative depth with respect to the camera, such as $\langle$human 3, in front of, human 2$\rangle$ and $\langle$human 2, in front of, human 1$\rangle$.
The ground truth relation $\langle$human\_1 [left arm], pull, human\_2$\rangle$ is  with low confidence and thus excluded from the top 25 relations.
In Figure~\ref{fig:result}b, SMS can predict the geometric relations that  human\_1 and human\_2 are next to each other and in front of building\_1. Moreover, $\langle$human\_1 [right shoe], kick, ball$\rangle$ is accurately predicted. Additionally, the human arms, legs and shoes are well segmented. 
In Figure~\ref{fig:result}c, the geometric graph contains the true positives including  $\langle$human\_1, next to, human\_2$\rangle$, $\langle$human\_1, stand on, beach\_1$\rangle$. 
Note that our method can estimate two action relations between a human part and another entity, such as 
$\langle$human\_1 [left arm], hold, human\_2$\rangle$ and $\langle$human\_1 [left arm], pull, human\_2$\rangle$.
In Figure~\ref{fig:result}d, due to occlusion by human\_3, human\_4 is not detected.
We can handle cases where two human parts of a person interact with another entity.
For example, we correctly predict    $\langle$human\_1 [right arm], touch, table\_1$\rangle$ and  $\langle$human\_1 [left arm], touch, table\_1$\rangle$ simultaneously. 
Another example is, we also successfully estimate both hands of human\_2 touching table\_1.
In Figure~\ref{fig:result}e,  the stuff categories such as ``water'' and ``beach'' are accurately segmented.
By referring to the  geometric graph including $\langle$human\_2, in front of, water\_1$\rangle$, 
$\langle$human\_2, lie on, beach\_1$\rangle$ and $\langle$human\_1, next to, human\_2$\rangle$, we can infer the relative position of human\_2 in the image.
The estimated action graph is very accurate, including  $\langle$human\_1 [right arm], touch, beach\_1$\rangle$  and $\langle$human\_1 [left arm], touch, beach\_1$\rangle$ etc.
In sum, the results show that SMS can well infer relations between human and surrounding entities, including things and stuff. Moreover, The interacted human parts can be accurately located. In addition, our method can describe entities by accurate segmentation masks.

\section{Conclusion} \label{sec:Conclusion}

In this work, we introduce the new HRS task,  \textcolor{black}{as a fine-grained case of HOI-det}.
HRS focuses on the relations between humans and surrounding entities including things and stuff, and utilizes the masks to represent subjects and objects. We also propose a new PIC dataset, which is a large scale HRS dataset with rich annotations including entity segmentation, relation and human parsing. Moreover, PIC is a high-resolution dataset which contains images several times larger than popular VRD and HOI-det datasets. We organized ECCV 2018 the 1st PIC challenge and ICCV 2019 the 2nd PIC  challenge using the initial versions of the PIC dataset. The challenges attracted many participating teams around the world. By integrating the feedbacks from these teams, we refined the PIC dataset to contribute the latest version in this paper. 
We also propose the SMS method to solve the HRS problem. SMS is able to generate segmentation results and relation detection results simultaneously and
output the HRS results in one pass. Experiments on PIC and V-COCO datasets show that SMS is more   efficient and effective than multiple two stage baselines. Dataset and code will be released. 
In further  we will explore  whether the HRS results can assist other high-level cognitive-like tasks, such as visual question answering and embodied AI. Moreover, we also plan to explore the video HRS problem which outputs the human entities relations and their spatial-temporal segmentation masks.

\IEEEdisplaynotcompsoctitleabstractindextext
	
\IEEEpeerreviewmaketitle


\bibliographystyle{IEEEtran}
\bibliography{reference}

\begin{thebibliography}{10}
\providecommand{\url}[1]{#1}
\csname url@samestyle\endcsname
\providecommand{\newblock}{\relax}
\providecommand{\bibinfo}[2]{#2}
\providecommand{\BIBentrySTDinterwordspacing}{\spaceskip=0pt\relax}
\providecommand{\BIBentryALTinterwordstretchfactor}{4}
\providecommand{\BIBentryALTinterwordspacing}{\spaceskip=\fontdimen2\font plus
\BIBentryALTinterwordstretchfactor\fontdimen3\font minus
  \fontdimen4\font\relax}
\providecommand{\BIBforeignlanguage}[2]{{%
\expandafter\ifx\csname l@#1\endcsname\relax
\typeout{** WARNING: IEEEtran.bst: No hyphenation pattern has been}%
\typeout{** loaded for the language `#1'. Using the pattern for}%
\typeout{** the default language instead.}%
\else
\language=\csname l@#1\endcsname
\fi
#2}}
\providecommand{\BIBdecl}{\relax}
\BIBdecl

\bibitem{habitat2020sim2real}
{Abhishek Kadian*}, {Joanne Truong*}, A.~Gokaslan, A.~Clegg, E.~Wijmans,
  S.~Lee, M.~Savva, S.~Chernova, and D.~Batra, ``Are {W}e {M}aking {R}eal
  {P}rogress in {S}imulated {E}nvironments? {M}easuring the {S}im2{R}eal {G}ap
  in {E}mbodied {V}isual {N}avigation,'' in \emph{arXiv:1912.06321}, 2019.

\bibitem{das2018embodied}
A.~Das, S.~Datta, G.~Gkioxari, S.~Lee, D.~Parikh, and D.~Batra, ``Embodied
  question answering,'' in \emph{Proceedings of the IEEE Conference on Computer
  Vision and Pattern Recognition Workshops}, 2018, pp. 2054--2063.

\bibitem{anderson2018vision}
P.~Anderson, Q.~Wu, D.~Teney, J.~Bruce, M.~Johnson, N.~S{\"u}nderhauf, I.~Reid,
  S.~Gould, and A.~van~den Hengel, ``Vision-and-language navigation:
  Interpreting visually-grounded navigation instructions in real
  environments,'' in \emph{Proceedings of the IEEE Conference on Computer
  Vision and Pattern Recognition}, 2018, pp. 3674--3683.

\bibitem{qi2020reverie}
Y.~Qi, Q.~Wu, P.~Anderson, X.~Wang, W.~Y. Wang, C.~Shen, and A.~v.~d. Hengel,
  ``Reverie: Remote embodied visual referring expression in real indoor
  environments,'' in \emph{Proceedings of the IEEE/CVF Conference on Computer
  Vision and Pattern Recognition}, 2020, pp. 9982--9991.

\bibitem{lu2019vilbert}
J.~Lu, D.~Batra, D.~Parikh, and S.~Lee, ``Vilbert: Pretraining task-agnostic
  visiolinguistic representations for vision-and-language tasks,'' in
  \emph{Advances in Neural Information Processing Systems}, 2019, pp. 13--23.

\bibitem{chen2019uniter}
Y.-C. Chen, L.~Li, L.~Yu, A.~E. Kholy, F.~Ahmed, Z.~Gan, Y.~Cheng, and J.~Liu,
  ``Uniter: Learning universal image-text representations,'' \emph{arXiv
  preprint arXiv:1909.11740}, 2019.

\bibitem{su2019vl}
W.~Su, X.~Zhu, Y.~Cao, B.~Li, L.~Lu, F.~Wei, and J.~Dai, ``Vl-bert:
  Pre-training of generic visual-linguistic representations,'' \emph{arXiv
  preprint arXiv:1908.08530}, 2019.

\bibitem{antol2015vqa}
S.~Antol, A.~Agrawal, J.~Lu, M.~Mitchell, D.~Batra, C.~Lawrence~Zitnick, and
  D.~Parikh, ``Vqa: Visual question answering,'' in \emph{Proceedings of the
  IEEE international conference on computer vision}, 2015, pp. 2425--2433.

\bibitem{anderson2018bottom}
P.~Anderson, X.~He, C.~Buehler, D.~Teney, M.~Johnson, S.~Gould, and L.~Zhang,
  ``Bottom-up and top-down attention for image captioning and visual question
  answering,'' in \emph{Proceedings of the IEEE conference on computer vision
  and pattern recognition}, 2018, pp. 6077--6086.

\bibitem{li2019manigan}
B.~Li, X.~Qi, T.~Lukasiewicz, and P.~H. Torr, ``Manigan: Text-guided image
  manipulation,'' \emph{arXiv preprint arXiv:1912.06203}, 2019.

\bibitem{krishna2017visual}
R.~Krishna, Y.~Zhu, O.~Groth, J.~Johnson, K.~Hata, J.~Kravitz, S.~Chen,
  Y.~Kalantidis, L.-J. Li, D.~A. Shamma \emph{et~al.}, ``Visual genome:
  Connecting language and vision using crowdsourced dense image annotations,''
  \emph{International Journal of Computer Vision}, vol. 123, no.~1, pp. 32--73,
  2017.

\bibitem{zellers2018neural}
R.~Zellers, M.~Yatskar, S.~Thomson, and Y.~Choi, ``Neural motifs: Scene graph
  parsing with global context,'' in \emph{Proceedings of the IEEE Conference on
  Computer Vision and Pattern Recognition}, 2018, pp. 5831--5840.

\bibitem{zhang2017visual}
H.~Zhang, Z.~Kyaw, S.-F. Chang, and T.-S. Chua, ``Visual translation embedding
  network for visual relation detection,'' in \emph{Proceedings of the IEEE
  conference on computer vision and pattern recognition}, 2017, pp. 5532--5540.

\bibitem{yang2018graph}
J.~Yang, J.~Lu, S.~Lee, D.~Batra, and D.~Parikh, ``Graph r-cnn for scene graph
  generation,'' in \emph{Proceedings of the European conference on computer
  vision}, 2018, pp. 670--685.

\bibitem{tang2020unbiased}
K.~Tang, Y.~Niu, J.~Huang, J.~Shi, and H.~Zhang, ``Unbiased scene graph
  generation from biased training,'' \emph{arXiv preprint arXiv:2002.11949},
  2020.

\bibitem{gupta2015visual}
S.~Gupta and J.~Malik, ``Visual semantic role labeling,'' \emph{arXiv preprint
  arXiv:1505.04474}, 2015.

\bibitem{chao2018learning}
Y.-W. Chao, Y.~Liu, X.~Liu, H.~Zeng, and J.~Deng, ``Learning to detect
  human-object interactions,'' in \emph{2018 ieee winter conference on
  applications of computer vision}.\hskip 1em plus 0.5em minus 0.4em\relax
  IEEE, 2018, pp. 381--389.

\bibitem{lu2016visual}
C.~Lu, R.~Krishna, M.~Bernstein, and L.~Fei-Fei, ``Visual relationship
  detection with language priors,'' in \emph{European Conference on Computer
  Vision}.\hskip 1em plus 0.5em minus 0.4em\relax Springer, 2016, pp. 852--869.

\bibitem{xu2017scene}
D.~Xu, Y.~Zhu, C.~B. Choy, and L.~Fei-Fei, ``Scene graph generation by
  iterative message passing,'' in \emph{Proceedings of the IEEE Conference on
  Computer Vision and Pattern Recognition}, vol.~2, 2017.

\bibitem{li2017vip}
Y.~Li, W.~Ouyang, X.~Wang, and X.~Tang, ``Vip-cnn: Visual phrase guided
  convolutional neural network,'' in \emph{Proceedings of the IEEE Conference
  on Computer Vision and Pattern Recognition}, 2017, pp. 1347--1356.

\bibitem{kipf2016semi}
T.~N. Kipf and M.~Welling, ``Semi-supervised classification with graph
  convolutional networks,'' \emph{arXiv preprint arXiv:1609.02907}, 2016.

\bibitem{wang2019deep}
T.~Wang, R.~M. Anwer, M.~H. Khan, F.~S. Khan, Y.~Pang, L.~Shao, and
  J.~Laaksonen, ``Deep contextual attention for human-object interaction
  detection,'' in \emph{Proceedings of the IEEE International Conference on
  Computer Vision}, 2019, pp. 5694--5702.

\bibitem{qi2018learning}
S.~Qi, W.~Wang, B.~Jia, J.~Shen, and S.-C. Zhu, ``Learning human-object
  interactions by graph parsing neural networks,'' in \emph{Proceedings of the
  European Conference on Computer Vision}, 2018, pp. 401--417.

\bibitem{feng2019turbo}
W.~Feng, W.~Liu, T.~Li, J.~Peng, C.~Qian, and X.~Hu, ``Turbo learning framework
  for human-object interactions recognition and human pose estimation,'' in
  \emph{Proceedings of the AAAI Conference on Artificial Intelligence},
  vol.~33, 2019, pp. 898--905.

\bibitem{wang2020learning}
T.~Wang, T.~Yang, M.~Danelljan, F.~S. Khan, X.~Zhang, and J.~Sun, ``Learning
  human-object interaction detection using interaction points,'' in
  \emph{Proceedings of the IEEE/CVF Conference on Computer Vision and Pattern
  Recognition}, 2020, pp. 4116--4125.

\bibitem{ren2016faster}
S.~Ren, K.~He, R.~Girshick, and J.~Sun, ``Faster r-cnn: Towards real-time
  object detection with region proposal networks,'' \emph{IEEE transactions on
  pattern analysis and machine intelligence}, vol.~39, no.~6, pp. 1137--1149,
  2016.

\bibitem{dai2016instance}
J.~Dai, K.~He, and J.~Sun, ``Instance-aware semantic segmentation via
  multi-task network cascades,'' in \emph{Proceedings of the IEEE Conference on
  Computer Vision and Pattern Recognition}, 2016, pp. 3150--3158.

\bibitem{liu2018path}
S.~Liu, L.~Qi, H.~Qin, J.~Shi, and J.~Jia, ``Path aggregation network for
  instance segmentation,'' in \emph{Proceedings of the IEEE Conference on
  Computer Vision and Pattern Recognition}, 2018, pp. 8759--8768.

\bibitem{li2017fully}
Y.~Li, H.~Qi, J.~Dai, X.~Ji, and Y.~Wei, ``Fully convolutional instance-aware
  semantic segmentation,'' in \emph{Proceedings of the IEEE Conference on
  Computer Vision and Pattern Recognition}, 2017, pp. 2359--2367.

\bibitem{he2017mask}
K.~He, G.~Gkioxari, P.~Doll{\'a}r, and R.~Girshick, ``Mask r-cnn,'' in
  \emph{Proceedings of the IEEE international conference on computer vision},
  2017, pp. 2961--2969.

\bibitem{chen2019hybrid}
K.~Chen, J.~Pang, J.~Wang, Y.~Xiong, X.~Li, S.~Sun, W.~Feng, Z.~Liu, J.~Shi,
  W.~Ouyang \emph{et~al.}, ``Hybrid task cascade for instance segmentation,''
  in \emph{Proceedings of the IEEE conference on computer vision and pattern
  recognition}, 2019, pp. 4974--4983.

\bibitem{fu2019retinamask}
C.-Y. Fu, M.~Shvets, and A.~C. Berg, ``Retinamask: Learning to predict masks
  improves state-of-the-art single-shot detection for free,'' \emph{arXiv
  preprint arXiv:1901.03353}, 2019.

\bibitem{newell2017associative}
A.~Newell, Z.~Huang, and J.~Deng, ``Associative embedding: End-to-end learning
  for joint detection and grouping,'' in \emph{Advances in neural information
  processing systems}, 2017, pp. 2277--2287.

\bibitem{fathi2017semantic}
A.~Fathi, Z.~Wojna, V.~Rathod, P.~Wang, H.~O. Song, S.~Guadarrama, and K.~P.
  Murphy, ``Semantic instance segmentation via deep metric learning,''
  \emph{arXiv preprint arXiv:1703.10277}, 2017.

\bibitem{neven2019instance}
D.~Neven, B.~D. Brabandere, M.~Proesmans, and L.~V. Gool, ``Instance
  segmentation by jointly optimizing spatial embeddings and clustering
  bandwidth,'' in \emph{Proceedings of the IEEE Conference on Computer Vision
  and Pattern Recognition}, 2019, pp. 8837--8845.

\bibitem{xie2019polarmask}
E.~Xie, P.~Sun, X.~Song, W.~Wang, X.~Liu, D.~Liang, C.~Shen, and P.~Luo,
  ``Polarmask: Single shot instance segmentation with polar representation,''
  \emph{arXiv preprint arXiv:1909.13226}, 2019.

\bibitem{wang2019solo}
X.~Wang, T.~Kong, C.~Shen, Y.~Jiang, and L.~Li, ``Solo: Segmenting objects by
  locations,'' \emph{arXiv preprint arXiv:1912.04488}, 2019.

\bibitem{wang2020solov2}
X.~Wang, R.~Zhang, T.~Kong, L.~Li, and C.~Shen, ``Solov2: Dynamic, faster and
  stronger,'' \emph{arXiv preprint arXiv:2003.10152}, 2020.

\bibitem{zhou2020cascaded}
T.~Zhou, W.~Wang, S.~Qi, H.~Ling, and J.~Shen, ``Cascaded human-object
  interaction recognition,'' in \emph{Proceedings of the IEEE/CVF Conference on
  Computer Vision and Pattern Recognition}, 2020, pp. 4263--4272.

\bibitem{chen2014detect}
X.~Chen, R.~Mottaghi, X.~Liu, S.~Fidler, R.~Urtasun, and A.~Yuille, ``Detect
  what you can: Detecting and representing objects using holistic models and
  body parts,'' in \emph{Proceedings of the IEEE Conference on Computer Vision
  and Pattern Recognition}, 2014, pp. 1971--1978.

\bibitem{liang2015deep}
X.~Liang, S.~Liu, X.~Shen, J.~Yang, L.~Liu, J.~Dong, L.~Lin, and S.~Yan, ``Deep
  human parsing with active template regression,'' \emph{IEEE transactions on
  pattern analysis and machine intelligence}, vol.~37, no.~12, pp. 2402--2414,
  2015.

\bibitem{liang2015human}
X.~Liang, C.~Xu, X.~Shen, J.~Yang, S.~Liu, J.~Tang, L.~Lin, and S.~Yan, ``Human
  parsing with contextualized convolutional neural network,'' in
  \emph{Proceedings of the IEEE international conference on computer vision},
  2015, pp. 1386--1394.

\bibitem{liang2018look}
X.~Liang, K.~Gong, X.~Shen, and L.~Lin, ``Look into person: Joint body parsing
  \& pose estimation network and a new benchmark,'' \emph{IEEE transactions on
  pattern analysis and machine intelligence}, vol.~41, no.~4, pp. 871--885,
  2018.

\bibitem{gong2018instance}
K.~Gong, X.~Liang, Y.~Li, Y.~Chen, M.~Yang, and L.~Lin, ``Instance-level human
  parsing via part grouping network,'' in \emph{Proceedings of the European
  Conference on Computer Vision}, 2018, pp. 770--785.

\bibitem{zhao2018understanding}
J.~Zhao, J.~Li, Y.~Cheng, T.~Sim, S.~Yan, and J.~Feng, ``Understanding humans
  in crowded scenes: Deep nested adversarial learning and a new benchmark for
  multi-human parsing,'' in \emph{Proceedings of the 26th ACM international
  conference on Multimedia}, 2018, pp. 792--800.

\bibitem{li2017multiple}
J.~Li, J.~Zhao, Y.~Wei, C.~Lang, Y.~Li, T.~Sim, S.~Yan, and J.~Feng,
  ``Multiple-human parsing in the wild,'' \emph{arXiv preprint
  arXiv:1705.07206}, 2017.

\bibitem{newell2016stacked}
A.~Newell, K.~Yang, and J.~Deng, ``Stacked hourglass networks for human pose
  estimation,'' in \emph{European conference on computer vision}.\hskip 1em
  plus 0.5em minus 0.4em\relax Springer, 2016, pp. 483--499.

\bibitem{law2018cornernet}
H.~Law and J.~Deng, ``Cornernet: Detecting objects as paired keypoints,'' in
  \emph{Proceedings of the European Conference on Computer Vision}, 2018, pp.
  734--750.

\bibitem{zhou2019objects}
X.~Zhou, D.~Wang, and P.~Kr{\"a}henb{\"u}hl, ``Objects as points,'' \emph{arXiv
  preprint arXiv:1904.07850}, 2019.

\bibitem{berman2018lovasz}
M.~Berman, A.~Rannen~Triki, and M.~B. Blaschko, ``The lov{\'a}sz-softmax loss:
  a tractable surrogate for the optimization of the intersection-over-union
  measure in neural networks,'' in \emph{Proceedings of the IEEE Conference on
  Computer Vision and Pattern Recognition}, 2018, pp. 4413--4421.

\bibitem{chen2019knowledge}
T.~Chen, W.~Yu, R.~Chen, and L.~Lin, ``Knowledge-embedded routing network for
  scene graph generation,'' in \emph{Proceedings of the IEEE Conference on
  Computer Vision and Pattern Recognition}, 2019, pp. 6163--6171.

\bibitem{wan2019pose}
B.~Wan, D.~Zhou, Y.~Liu, R.~Li, and X.~He, ``Pose-aware multi-level feature
  network for human object interaction detection,'' in \emph{Proceedings of the
  IEEE International Conference on Computer Vision}, 2019, pp. 9469--9478.

\bibitem{yu2018deep}
F.~Yu, D.~Wang, E.~Shelhamer, and T.~Darrell, ``Deep layer aggregation,'' in
  \emph{Proceedings of the IEEE conference on computer vision and pattern
  recognition}, 2018, pp. 2403--2412.

\bibitem{he2016deep}
K.~He, X.~Zhang, S.~Ren, and J.~Sun, ``Deep residual learning for image
  recognition,'' in \emph{Proceedings of the IEEE conference on computer vision
  and pattern recognition}, 2016, pp. 770--778.

\bibitem{lin2017feature}
T.-Y. Lin, P.~Doll{\'a}r, R.~Girshick, K.~He, B.~Hariharan, and S.~Belongie,
  ``Feature pyramid networks for object detection,'' in \emph{Proceedings of
  the IEEE conference on computer vision and pattern recognition}, 2017, pp.
  2117--2125.

\bibitem{kirillov2019panoptic}
A.~Kirillov, R.~Girshick, K.~He, and P.~Doll{\'a}r, ``Panoptic feature pyramid
  networks,'' in \emph{Proceedings of the IEEE Conference on Computer Vision
  and Pattern Recognition}, 2019, pp. 6399--6408.

\bibitem{deng2009imagenet}
J.~Deng, W.~Dong, R.~Socher, L.-J. Li, K.~Li, and L.~Fei-Fei, ``Imagenet: A
  large-scale hierarchical image database,'' in \emph{2009 IEEE conference on
  computer vision and pattern recognition}.\hskip 1em plus 0.5em minus
  0.4em\relax Ieee, 2009, pp. 248--255.

\bibitem{kingma2014adam}
D.~P. Kingma and J.~Ba, ``Adam: A method for stochastic optimization,''
  \emph{arXiv preprint arXiv:1412.6980}, 2014.

\bibitem{ren2015faster}
S.~Ren, K.~He, R.~Girshick, and J.~Sun, ``Faster r-cnn: Towards real-time
  object detection with region proposal networks,'' in \emph{Advances in neural
  information processing systems}, 2015, pp. 91--99.

\bibitem{misra2016seeing}
I.~Misra, C.~Lawrence~Zitnick, M.~Mitchell, and R.~Girshick, ``Seeing through
  the human reporting bias: Visual classifiers from noisy human-centric
  labels,'' in \emph{Proceedings of the IEEE Conference on Computer Vision and
  Pattern Recognition}, 2016, pp. 2930--2939.

\bibitem{zhu2019deformable}
X.~Zhu, H.~Hu, S.~Lin, and J.~Dai, ``Deformable convnets v2: More deformable,
  better results,'' in \emph{Proceedings of the IEEE Conference on Computer
  Vision and Pattern Recognition}, 2019, pp. 9308--9316.

\bibitem{xie2020polarmask}
E.~Xie, P.~Sun, X.~Song, W.~Wang, X.~Liu, D.~Liang, C.~Shen, and P.~Luo,
  ``Polarmask: Single shot instance segmentation with polar representation,''
  in \emph{Proceedings of the IEEE/CVF Conference on Computer Vision and
  Pattern Recognition}, 2020, pp. 12\,193--12\,202.

\end{thebibliography}

\end{document}